\documentclass[sigconf, 9pt]{acmart}

\usepackage[english]{babel}
\usepackage{blindtext}
\usepackage{caption}
\usepackage{subcaption}
\usepackage{color}
\usepackage{tabularx}
\usepackage{multirow}
\usepackage{multicol}
\usepackage{graphicx}
\usepackage{tabularray}
\usepackage{float}
\usepackage{xcolor} 
\usepackage{todonotes}
\usepackage{array} 
\usepackage{makecell} 
\usepackage{booktabs} 
\usepackage{algorithm}
\usepackage{algorithmic}

\setcopyright{acmcopyright}

\newcommand{\name}{MMEdge\xspace}

\acmYear{2026}\copyrightyear{2026}
\setcopyright{cc}
\setcctype{by}
\acmConference[SenSys '26]{ACM/IEEE International Conference on Embedded Artificial Intelligence and Sensing Systems}{May 11--14, 2026}{Saint Malo, France}
\acmBooktitle{ACM/IEEE International Conference on Embedded Artificial Intelligence and Sensing Systems (SenSys '26), May 11--14, 2026, Saint Malo, France}
\acmDOI{10.1145/3774906.3800485}
\acmISBN{979-8-4007-2309-4/26/05}

\acmPrice{}

\begin{document}

\title{MMEdge: Accelerating On-device Multimodal Inference via Pipelined Sensing and Encoding}

\author{Runxi Huang, Mingxuan Yu, Mingyu Tsoi, Xiaomin Ouyang}
\authornote{Corresponding author}
\affiliation{
  \institution{Hong Kong University of Science and Technology}
  \city{Hong Kong SAR}
  \country{China}
}
\email{rhuangbj@connect.ust.hk, myuau@connect.ust.hk, mytsoi@connect.ust.hk, xmouyang@cse.ust.hk}


\begin{abstract}

Real-time multimodal inference on resource-constrained edge devices is essential for applications such as autonomous driving, human-computer interaction, and mobile health. However, prior work often overlooks the tight coupling between sensing dynamics and model execution, as well as the complex inter-modality dependencies.
In this paper, we propose MMEdge, a new on-device multimodal inference framework based on \textit{pipelined sensing and encoding}. Instead of waiting for complete sensor inputs, MMEdge decomposes the entire inference process into a sequence of fine-grained sensing and encoding units, allowing computation to proceed incrementally as data arrive. 
MMEdge also introduces a lightweight but effective temporal aggregation module that captures rich temporal dynamics across different pipelined units to maintain accuracy performance.
Such pipelined design also opens up opportunities for fine-grained cross-modal optimization and early decision-making during inference. 
To further enhance system performance under resource variability and input data complexity, MMEdge incorporates an \emph{adaptive multimodal configuration optimizer} that dynamically selects optimal sensing and model configurations for each modality under latency constraints, and a \emph{cross-modal speculative skipping} mechanism that bypasses future units of slower modalities when early predictions reach sufficient confidence.
We evaluate MMEdge using two public multimodal datasets and deploy it on a real-world unmanned aerial vehicle (UAV)-based multimodal testbed. The results show that MMEdge significantly reduces end-to-end latency while maintaining high task accuracy across various system and data dynamics. A video demonstration of \name's performance in real world is available at \href{https://youtu.be/qRew7sT-iWw}{https://youtu.be/qRew7sT-iWw}.  \footnote{Code available at: \href{https://github.com/HKUST-MINSys-Lab/MMEdge}{https://github.com/HKUST-MINSys-Lab/MMEdge}.}

\end{abstract}


\keywords{Multimodal Sensing Systems, On-device AI, Efficient ML}

\ccsdesc[500]{Computing methodologies~Artificial intelligence}
\ccsdesc[500]{Computing methodologies~Machine learning}
\ccsdesc[500]{Computer systems organization~Embedded and cyber-physical systems}

\maketitle

\section{Introduction}

Multimodal sensing and learning systems are increasingly adopted in real-time applications such as autonomous driving \cite{xie2023timely, wu2024adaflow}, fall detection \cite{chen2020fall}, and wearable interaction \cite{mayberry2015cider, pang2023ubipose}. Such systems continuously collect, process and fuse data from heterogeneous sensors (e.g., cameras, microphones, IMUs) to provide improved performance for complex tasks in dynamic environments. For example, RCBEVDet~\cite{lin2024rcbevdet} shows that integrating radar with camera can significantly enhance object detection accuracy in autonomous driving. While offloading computation to the cloud has been widely adopted in distributed multimodal sensing systems, it raises growing concerns about user's data privacy. Moreover, such systems often experience significant latency due to unpredictable network dynamics \cite{li2021low, huang2023re} and the high bandwidth demands of data-intensive sensors like LiDAR \cite{wu2024adaflow}. As a result, there is an increasing demand for efficient on-device multimodal systems, which enables end-to-end processing of sensor data locally without transmitting sensor data.

However, deploying end-to-end multimodal perception systems on devices presents several major challenges, particularly under limited and dynamic resource conditions. First, there is an inherent dependency between the sensing and model inference phases when they share computational resources on the same device, making isolated optimization strategies ineffective. For instance, increasing model complexity or enhancing sensing granularity can both improve prediction accuracy, but also introduce higher latency, necessitating careful coordination between the two. Second, multimodal systems also exhibit strong inter-modality dependencies, where both fusion accuracy and latency depend on the joint optimization of all modalities. For example, Shuai et al.~\cite{shuai2021millieye} show that human detection accuracy in radar-camera systems depends on synchronized multimodal inputs, especially under varying lighting conditions, rather than relying on any single modality alone. Moreover, there will be asynchronous processing delays across modalities, which often leads to idle waiting time and inefficient resource utilization. In such cases, faster modalities are forced to wait for slower ones, increasing end-to-end latency. These challenges motivate a unified system design that dynamically coordinates the sensing and inference configurations across different modalities at run time to optimize both accuracy and efficiency.

Unfortunately, most prior work on on-device multimodal systems has focused on optimizing either the sensing or the model inference stage in isolation, without considering their interdependencies. Some methods focus on mitigating delays caused by asynchronous sensor inputs. For example, \cite{li2021low} introduced modality imputation to synthesize missing data from slow modalities using fast ones to reduce waiting time, while modality gating methods \cite{hou2023smg, mayberry2015cider} selectively skip uninformative sensor modalities before inference to improve efficiency. On the other hand, several methods aim at optimizing the model inference process itself through early exit strategies \cite{cao2022mobivqa} or expert model selection \cite{liu2024deepseek}. However, our results indicate that optimizing sensing or inference in isolation performs poorly under real-world system and data dynamics. Although some distributed systems consider sensing-inference interactions \cite{li2021low, wu2024adaflow}, they are designed to mitigate the impact of network dynamics during sensor data transmission \cite{rastikerdar2024cactus, huang2023re}, which cannot address the shared computational resource constraints between sensing and inference that are intrinsic to on-device multimodal systems. 

In this paper, we propose MMEdge, a new and efficient on-device multimodal inference system based on pipelined sensing and encoding.
Traditional end-to-end multimodal systems typically operate in a sequential manner, where model inference is blocked until all sensor data within a time window are fully acquired. Such sequential execution not only leads to accumulated and imbalanced delays across different modalities, but also incurs significant memory overhead at runtime. MMEdge addresses these limitations by decomposing the entire computation task into a sequence of fine-grained sensing and encoding units, each unit corresponding to the smallest data segment (e.g., a video frame or audio chunk). 
These units are processed immediately upon arrival, enabling feature encoding to occur during the sensing interval. As a result, such design eliminates idle periods between sensing intervals, and allow fully pipelined execution of data acquisition and feature encoding without waiting for complete time-window data.
To mitigate potential performance degradation from this decoupled processing, MMEdge introduces a lightweight temporal aggregation module that selectively preserve dependencies and contextual continuity across units. Specifically, this module exploits alternating temporal shift operations across units and extracts multi-scale temporal difference features to capture short- and long-term temporal correlations. This design ensures that critical temporal and semantic relationships are retained, even when data is processed in a fine-grained, pipelined manner. Moreover, to adapt to runtime resource variability and input data complexity, MMEdge incorporates an adaptive multimodal configuration optimizer that dynamically selects optimal sensing and model configurations for each modality to satisfy latency constraints under varying system conditions; and a cross-modal speculative skipping mechanism that reduces waiting time by bypassing future units of slower modalities when early predictions achieve sufficient confidence.

We deploy MMEdge on a real-world multimodal sensor testbed on Unmanned Aerial Vehicles (UAVs) for real-time human tracking tasks. The results show that MMedge significantly reduces end-to-end latency by 75.83\% without compromising task performance. We also evaluate the performance of MMEdge on Nvidia edge devices using other two public multimodal datasets. Our extensive evaluations show that MMEdge reduces inference latency while maintaining high task accuracy across diverse runtime conditions.

The main contributions of this work are:

\begin{itemize}
    \item  We conduct an in-depth analysis and evaluations of end-to-end latency in on-device multimodal systems to identify the key challenges, and show the potential of decomposing inference tasks into fine-grained units to accelerate processing.
    \item We propose MMEdge, a new on-device multimodal inference framework that decomposes the entire inference task into fine-grained sensing and encoding units for pipelined processing to effectively reduce end-to-end latency, and integrates a lightweight temporal aggregation module to capture temporal continuity across units to maintain accuracy.
    \item To further adapt to runtime data and system dynamics, MMEdge incorporates an adaptive multimodal configuration optimizer that adaptively adjusts sensing and model configurations for each modality, and employs a cross-modal speculative skipping module to reduce waiting time by bypassing data of slower modalities.
    \item We evaluate MMEdge using two public multimodal datasets and on a real-world UAV testbed. Our evaluation shows that MMEdge significantly reduces end-to-end latency while maintaining accuracy under dynamic runtime conditions.
\end{itemize}


\section{Related Work}
\label{sec:related work}

\noindent\textbf{Multimodal sensing and learning systems}. 
Multimodal sensing and learning systems are becoming increasingly prevalent in real-world applications. These systems continuously collect and fuse data from different sensor modalities to enhance performance in complex, dynamic environments. For instance, ADMarker \cite{ouyang2024admarker} fuses depth, radar, and audio data to detect daily activities for monitoring digital biomarkers of Alzheimer’s Disease. Similarly, autonomous driving~\cite{shi2022vips,he2023vi} and robotic~\cite{gtsam, kummerle2011g, wang2023pypose} systems leverage multiple sensor modalities to expand sensing coverage and maintain robust performance under varying environmental conditions.
However, most work in this space focuses on improving the accuracy of multimodal fusion, often overlooking inference latency, which is a critical factor for time-sensitive applications in practical deployments.

 \noindent\textbf{Accelerating on-device inference}. 
Reducing on-device inference latency is crucial for deploying deep models on resource-constrained devices.
Prior works explore dynamic adaptation strategies, including early-exit mechanisms~\cite{leontiadis2021s, jeon2023harvnet}, model or input reconfiguration~\cite{rastikerdar2024cactus, lee2024panopticus, mayberry2015cider, yang2020resolution, lu2010jigsaw, naderiparizi2017glimpse}, and parallel execution frameworks~\cite{wei2023nn, ling2022blastnet}.
For instance, CACTUS~\cite{rastikerdar2024cactus} selects micro-classifiers based on context, while Glimpse~\cite{naderiparizi2017glimpse} prunes irrelevant inputs using auxiliary sensors.
However, these methods focus on single-modality inference and neglect the end-to-end coupling between sensing and computation.
In contrast, MMEdge introduces \emph{pipelined sensing and encoding} for multimodal systems, bridging sensor data acquisition and model execution in a latency-aware manner.

\noindent \textbf{Efficient multimodal inference.}
Achieving real-time multimodal sensing and inference on devices is challenging due to the overhead of processing multiple modalities. Some works \cite{mayberry2015cider, cao2022mobivqa, xie2023timely} leverage prior knowledge from one modality to guide inference in another. For example, Glimpse~\cite{naderiparizi2017glimpse} uses auxiliary sensors to filter out irrelevant modalities, while CIDER~\cite{mayberry2015cider} employs infrared sensors to dynamically switch vision model configurations. Other approaches aim to achieve adaptive multimodal fusion during inference. MobiVQA~\cite{cao2022mobivqa} adjusts visual processing branches based on textual information, and SMG~\cite{hou2023smg} introduces a lightweight fusion method that selects informative features from both modalities for efficient inference.
However, most of these works focus on optimizing either the sensing or inference stage in isolation. Although some distributed systems consider sensing-inference interactions~\cite{li2021low, wu2024adaflow}, they are designed to mitigate the impact of network dynamics during sensor data transmission~\cite{huang2023re, rastikerdar2024cactus}, which cannot address the shared computational resource constraints between sensing and inference that are intrinsic to on-device multimodal systems.



\section{A Motivation Study}
\label{sec:motivate_study}

In this section, we evaluate the end-to-end latency of on-device multimodal inference systems and explore the potential of pipelined sensing and encoding to enhance processing efficiency.

\subsection{Understanding On-Device Multimodal Systems}

Achieving real-time multimodal sensing and inference on devices is important for applications that need reside data locally due to privacy concerns and require timely predictions. These systems typically perform end-to-end processing of synchronized multimodal streaming data within a time window to provide accurate perception and decision-making. For example, autonomous vehicles continuously collect data from cameras and mmWave radar, and must fuse this synchronized multimodal data for perception \cite{xie2023timely} and planning \cite{hu2023planning}, often within 100 ms to enable timely control decisions \cite{luo2019time}. Another example is interactive systems that rely on audio-visual streams for tasks such as speech recognition \cite{petridis2018end} and gesture recognition \cite{shi2021face}. Enabling low-latency multimodal sensing and inference directly on devices is essential for delivering responsive and preserving user data privacy.



\begin{figure}
    \centering
    \setlength{\abovecaptionskip}{0.cm}
    \setlength{\belowcaptionskip}{0.cm}
    \includegraphics[width=1.0\linewidth]{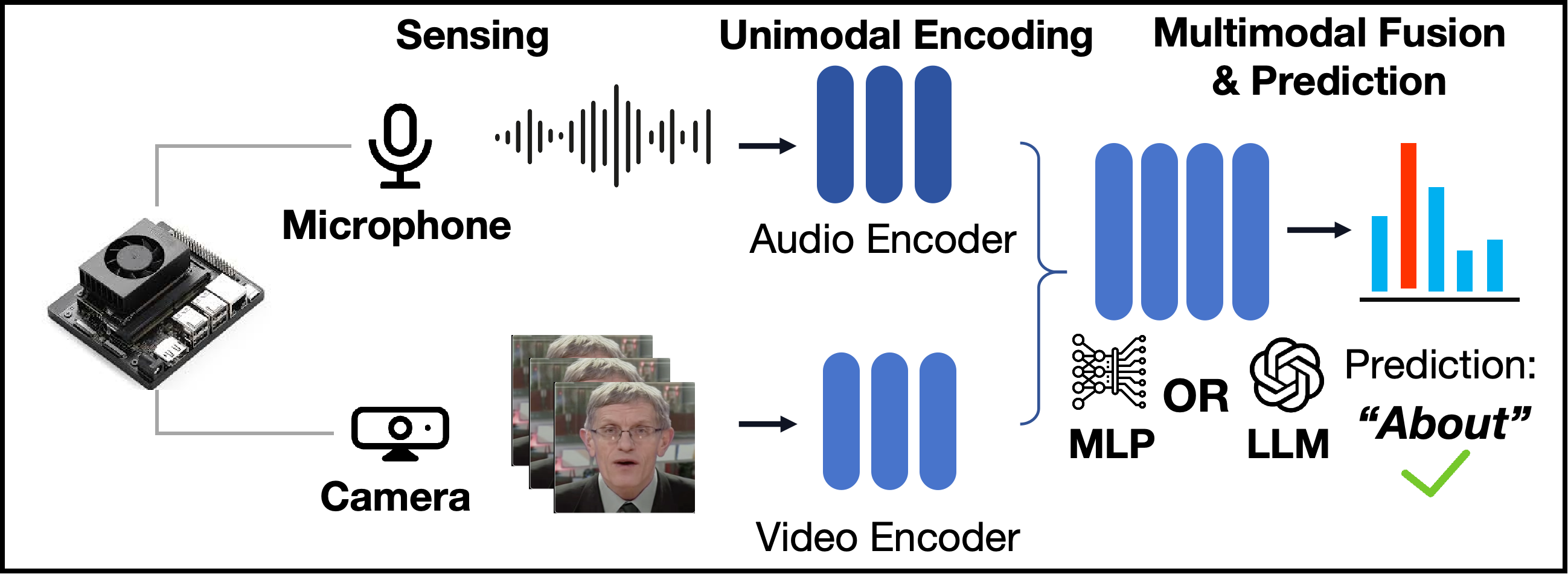}
    \caption{An end-to-end multimodal inference system that encompasses the entire data lifecycle on devices.
    }
    \label{fig:end_to_end}
    \vspace{-2em}
\end{figure}

\textbf{Traditional end-to-end inference pipeline.} Figure~\ref{fig:end_to_end} illustrates a typical on-device multimodal inference system, comprising the following stages:
(1) \emph{Data Collection and Preprocessing.} Sensors (e.g., cameras, microphones) capture raw physical signals. These signals are then preprocessed through operations like normalization, rescaling, and cropping to ensure compatibility with downstream deep learning models.
(2) \emph{Unimodal Encoding.} Preprocessed data from each modality is passed through different unimodal encoders (e.g., ResNet~\cite{he2016deep}, Transformer~\cite{vaswani2017attention}) to extract compact and informative unimodal feature embeddings.
(3) \emph{Multimodal Fusion.} The unimodal embeddings are fused via concatenation, attention mechanisms, or transformer-based fusion, to form a unified multimodal representation that aligns features across modalities.
(4) \emph{Prediction.} The fused representation is then used for task-specific inference, such as classification, detection, or generation, typically through MLP layers or decoders in multimodal models.

\begin{figure}
    \raggedright   
    \setlength{\abovecaptionskip}{0.cm}
    \setlength{\belowcaptionskip}{-0.cm}

    \begin{subfigure}{1\linewidth}
        \centering
        \includegraphics[width=\textwidth]{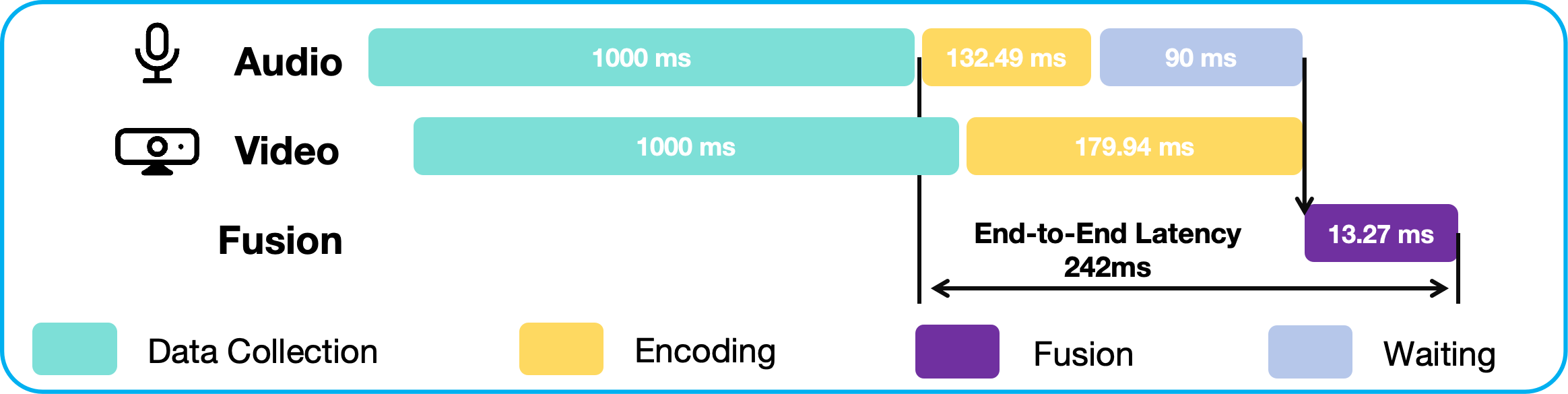}
        \caption{Traditional framework (Latency: 242ms|Accuracy: 92.76\%).}
        \label{fig:entire inference}
    \end{subfigure}
    \begin{subfigure}{1\linewidth}
        \centering
        \includegraphics[width=\textwidth]{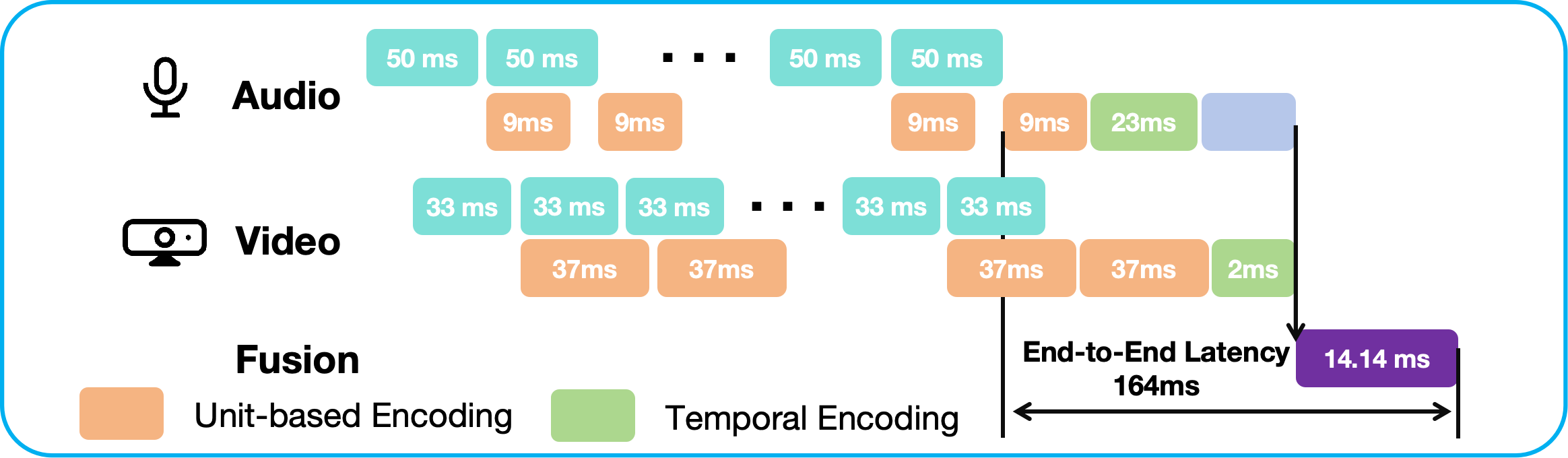}
        \caption{The new pipelined framework (Latency: 164ms|Accuracy:  72.44\%).}
        \label{fig:pipelined inference}
    \end{subfigure}

    \caption{Comparison between two inference frameworks.
    } 
    \label{fig: disentangling}
    \vspace{-2em}
\end{figure}

\textbf{Challenges.}
We conduct a motivational study to examine the challenges of optimizing inference latency in on-device multimodal systems. Specifically, we evaluate an audio-visual speech recognition task using the Lip Reading in the Wild dataset~\cite{chung2017lip}. The experiments are conducted on an NVIDIA Jetson Xavier NX (16GB memory), configured in 2-core, 10W power mode to simulate a low-resource deployment environment. We use ResNet-50 for video backbone and a 2-layer CNN for audio encoder, and an attention-based module for multimodal fusion. Each sample consists of a 1-second recording, comprising video at 30 FPS, and audio at 16 kHz segmented into chunks of 800 samples. We simulate data collection by loading data at fixed intervals matching the original sensor frame rate. A background thread continuously collects data to emulate realistic sensing overheads. Figure~\ref{fig: disentangling} shows the measured latency across different stages of the end-to-end inference process.

The results reveal two key challenges in accelerating on-device multimodal inference systems. First, there are imbalanced processing delays across different modalities, which prolongs overall inference latency as faster modalities need to wait for slower ones before fusion. For example, due to the larger data volume and model complexity, video processing takes significantly longer than audio data for both data collection and unimodal encoding, resulting in an idle waiting time of approximately 100 ms before fusion. Second, existing systems follow a sequential execution framework, where inference is blocked until all sensor data within a time window is fully acquired. This sequential dependency further amplifies end-to-end latency, as delays accumulate across modalities and samples. As shown in Fig.~\ref{fig:entire inference}, a 90 ms delay in the first sample causes all subsequent samples to be postponed, leading to a cumulative latency increase over time.
These challenges highlight the need for a unified system design that dynamically coordinates sensing and model inference across modalities at runtime for optimizing efficiency and accuracy of on-device multimodal systems.

\subsection{Potential of Pipelined Sensing and Encoding}
\label{sec:motivation_pipeline}
To address the limitations of traditional sequential processing, we explore decomposing the inference task into a pipelined sensing and encoding framework. As illustrated in Figure~\ref{fig:pipelined inference}, without waiting for data of the complete time window, we segment the input data into smaller units (e.g., individual video frames or audio chunks) and begin encoding each unit immediately upon arrival. Once the full sequence is collected, the unimodal features are aggregated across units for fusion and prediction.
Such framework allows encoding to proceed concurrently during the sensing interval. 

\textbf{Comparison between the traditional and pipelined framework.} We implement the pipelined inference framework using the same experimental setup as the traditional framework described in Section 3.1 on the audio-visual speech recognition task. For a fair comparison, both frameworks use the same backbone models. In the traditional setup, we use ResNet-50~\cite{he2016deep} for video encoding and a 2-layer CNN for audio encoding. In the pipelined framework, since each unit contains less temporal context (e.g., only one frame for video modality), we adopt a 2D ResNet without temporal modeling to better suit the finer granularity of input.

As shown in Figure~\ref{fig:entire inference}, the traditional pipeline waits for all sensor data within a time window to be collected before processing begins. This leads to high latency due to underutilization of the sensing interval, as well as asynchronous arrival and processing of audio and video data. In contrast, the pipelined framework reduces end-to-end latency by decomposing the unimodal encoding process and overlapping it with the sensing interval, allowing encoding to begin during data acquisition.
As shown in Fig~\ref{fig:pipelined inference}, pipelined execution improves efficiency in both unimodal encoding and overall inference latency, reducing latency by approximately 80 ms compared to the traditional pipeline. However, this latency reduction comes at the cost of accuracy. This is because, the global temporal modeling at the data level is replaced by aggregation across feature extracted on smaller units, which limits access to holistic temporal context. This results in an approximate 20\% drop in accuracy compared to the traditional pipeline.
Therefore, this requires careful design to effectively aggregate features across units and mitigate the loss of temporal coherence.

\begin{figure}
    \raggedright   
    \setlength{\abovecaptionskip}{0.cm}
    \setlength{\belowcaptionskip}{-0.cm}
    \begin{subfigure}{\linewidth}
        \centering
        \setlength{\abovecaptionskip}{0.cm}
        \setlength{\belowcaptionskip}{0.cm}
        \includegraphics[width=\linewidth]{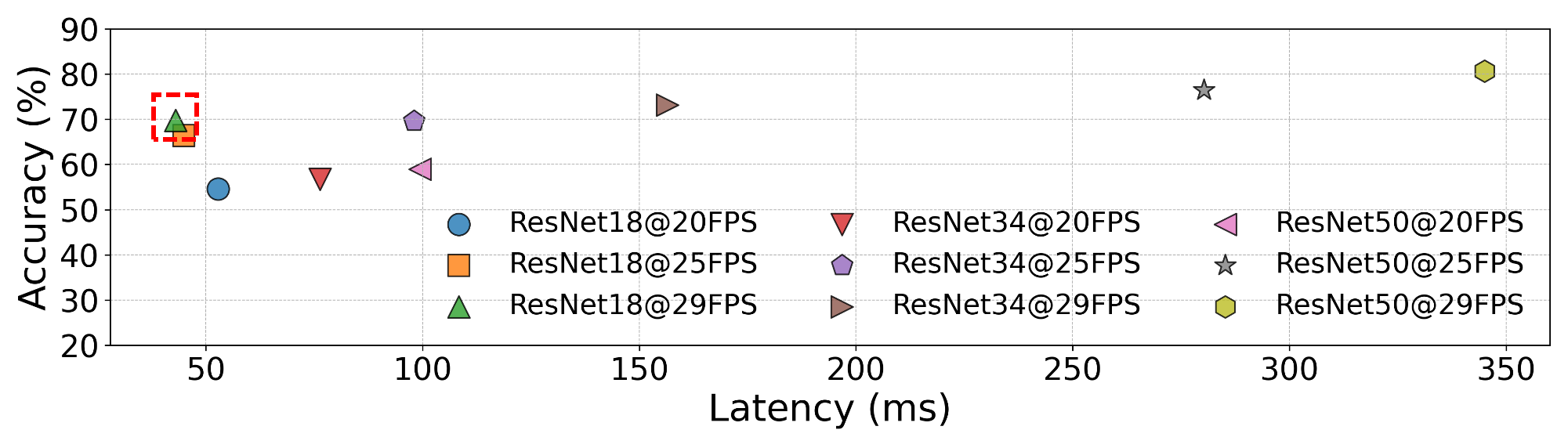}
        \caption{Dependencies between sensing and model configuration on video data (fixed configurations for audio).}
        \label{fig:sensing model dependency}
    \end{subfigure}
    \hspace{2mm}
    \begin{subfigure}{\linewidth}
        \centering
        \setlength{\abovecaptionskip}{0.cm}
        \setlength{\belowcaptionskip}{0.cm}
        \includegraphics[width=1\linewidth]{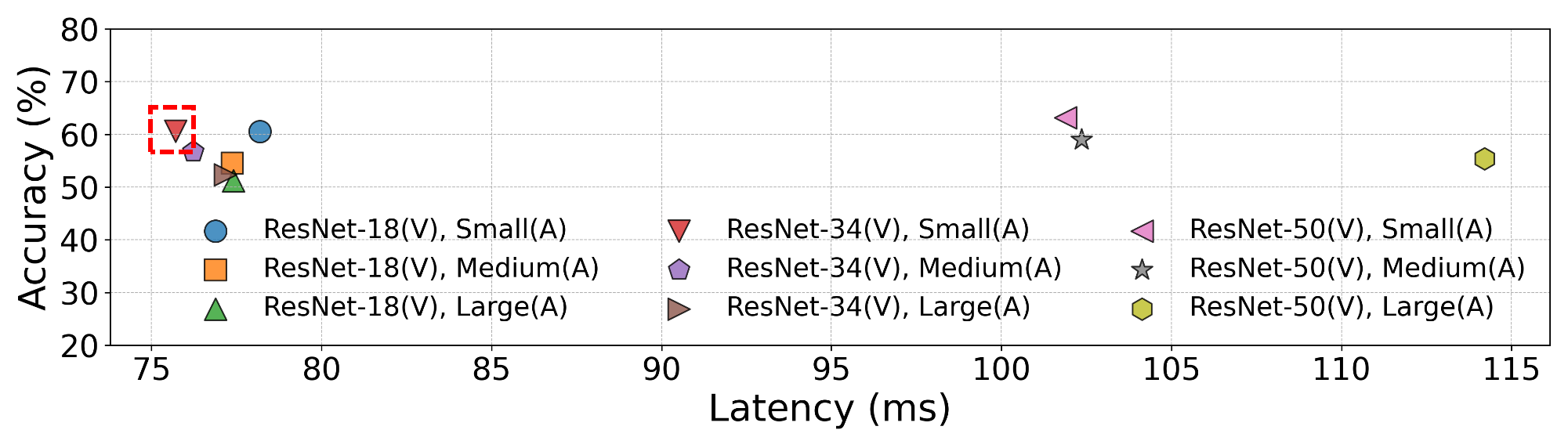}
        \caption{Dependencies of model configuration across modalities (fixed sensing configurations). 
        }
        \label{fig:multimodal dependency}
    \end{subfigure}
    \caption{Impact of different pipelined configurations. The upper-left region indicates better strategies that achieve higher accuracy with lower latency. 
    }
    \label{fig: dependencies}
    \vspace{-2em}
\end{figure}

\textbf{Impact of different pipelined configurations.} In the pipelined framework, different choices of sensing units and model complexities can affect latency and accuracy of multimodal inference systems.
To better understand how different configureations affect system performance, we profile 81 multimodal configurations by varying model sizes and sensing parameters. Specifically, we consider three video models (ResNet-18/34/50), three frame rates (20/25/29 FPS) for videos (50/40/33 ms for each unit), three audio models (Small/Medium/Large), and three audio chunk durations (50/62.5/75 ms for each unit). For each configuration, we measure both end-to-end latency on edge devices and top-1 accuracy.

Figure~\ref{fig:sensing model dependency} shows the accuracy and latency performance when varying model complexity and frame rate for video, while keeping the audio modality configuration fixed. 
The latency is measured as the end-to-end system delay, which accounts for both sensing and model inference across multiple modalities.
The upper-left region of the plot indicates strategies that achieve high accuracy with low latency, which is more favorable strategies for real-time applications. The results reveal inherent dependencies between sensing parameters and model configuration: increasing model depth and frame rate does not always lead to proportional gains in accuracy and may introduce unnecessary latency. For example, the accuracy of ResNet-34 with 25 FPS is comparable to ResNet-50 with 20 FPS but with significantly lower latency. Therefore, this highlights the importance of carefully selecting both the model and data segmentation strategy to achieve an optimal balance. For instance, if the target accuracy is 85\%, the configuration using ResNet-34 at 25 FPS delivers high accuracy with minimal latency.
Figure~\ref{fig:multimodal dependency} explores the impact of jointly selecting audio and video model configurations, while keeping their sensing parameters fixed. The results show that performance improves more significantly when both modalities are scaled together, indicating strong cross-modal complementarity. For example, replacing ResNet-18 with ResNet-34 while reducing the audio model from medium to small improves both accuracy and latency. This suggests complex cross-modal dependencies, where jointly adjusting configurations may yield better trade-offs than scaling a single modality alone.
These findings emphasize the importance of joint optimization of sensing and model configurations across modalities to meet real-time constraints while maintaining acceptable accuracy.

\subsection{Summary}

We now summarize the key findings from our motivation study.

\begin{itemize}
    \item On-device multimodal systems are subject to considerable latency during both the sensing and inference stages. Conventional sequential processing frameworks introduce imbalanced and accumulated delays across different modalities, prolonging the overall system delay.

    \item Pipelined sensing and encoding can reduce end-to-end latency, but also leads to degradation in accuracy performance. Moreover, significant dependencies exist between sensing and model inference stages, as well as across different modalities, requiring careful joint system optimization.
\end{itemize}

\begin{figure*}
    \centering
    \setlength{\abovecaptionskip}{0.cm}
    \setlength{\belowcaptionskip}{-0.cm}
    \includegraphics[width=\textwidth]{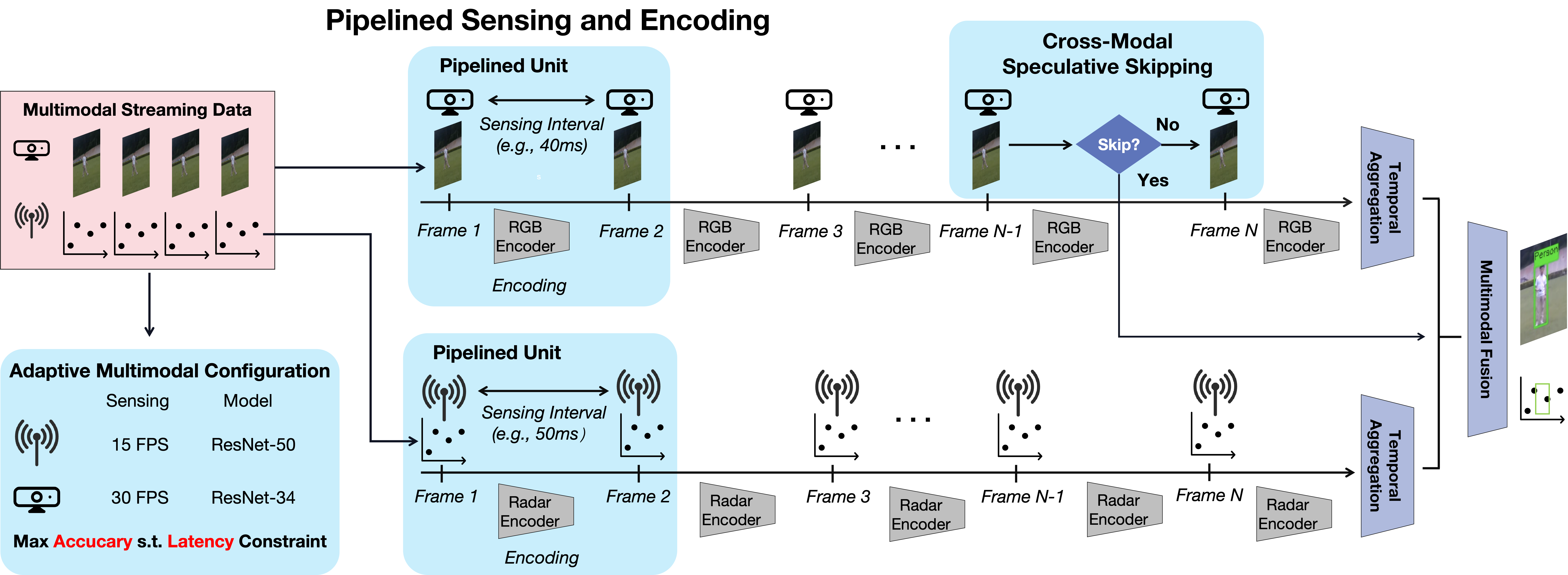}
    \caption{System Overview of MMEdge. MMEdge features a new pipelined sensing and encoding framework that decomposes the entire inference task into fine-grained units for paralell execution. It also integrates an adaptive multimodal configuration module that selects sensing and model configuration for each modality adapting to varying inputs and resource dynamics; and a cross-modal speculative skipping module to selectively skip the slower modalities.  
    }
    
    \label{fig:system overview}
    \vspace{-1em}
\end{figure*}

\section{System Overview}

We propose MMEdge, a real-time end-to-end multimodal inference system for temporal tasks under strict latency constraints, as illustrated in Figure~\ref{fig:system overview}. Traditional multimodal systems typically adopt a sequential pipeline, where inference is blocked until all modality inputs within a time window are collected, resulting in accumulated delays and increased memory usage. To overcome these limitations, MMEdge decomposes the entire task into a sequence of fine-grained sensing and encoding units, each aligned with a sensing interval (e.g., a video frame or an audio chunk).
These units are immediately processed upon arrival by lightweight encoders, enabling pipelined execution that overlaps sensing with inference and avoids idle waiting for full-window inputs.

Specifically, MMEdge decomposes inference into a sequence of fine-grained pipelined units, where each unit corresponds to the minimal data segment collected by a sensor at each sensing interval (e.g., a video frame or audio chunk), and is immediately passed through a lightweight encoder upon arrival. This design enables parallel sensing and inference, leveraging idle time between sensing intervals for unimodal feature encoding. Such a design also enables feature encoding of earlier segments to overlap with ongoing sensing and reduces system memory usage.
To mitigate potential performance degradation caused by broken temporal dependencies across units, we introduce a lightweight temporal aggregation module. This module applies alternating temporal shift operations and multi-scale temporal difference feature extraction to capture both short- and long-term temporal correlations across units. 



To adapt to runtime resource variability and input data complexities, MMEdge employs an adaptive multimodal configuration optimizer that dynamically selects the sensing and model configurations for each modality at runtime based on system and data dynamics. The optimization variables include granularity of pipelined processing units such as frame rate and audio chunk size, as well as model choices such as encoders with different model sizes. MMEdge also features an accuracy predictor that is trained offline to estimate the impact of different configurations on prediction accuracy. To reduce the computation overhead of running the accuracy predictor, we extract lightweight indicators—modality consistency and complementarity—from the original multimodal sensor data to guide the selection of appropriate configurations. The optimizer selects the configuration with the highest predicted accuracy that meets the given latency constraint.

Moreover, to address the imbalanced delays across different modalities, MMEdge incorporates a cross-modal speculative skipping strategy that reduces waiting time by bypassing slow modalities when early predictions achieve sufficient confidence. When data from faster modalities (e.g., audio) provides sufficient information for a confident prediction, the system can selectively skip inference of slower modalities (e.g., video) to reduce redundant computation.
Based on this design, MMEdge can adaptively partition, schedule, and execute multimodal inference in real time, and jointly optimize end-to-end stages across different modalities to achieve low-latency and accurate multimodal inference on resource-constrained devices.

\section{Design of MMEdge}
\label{sec:design of mmedge}

The design of MMEdge is motivated by the key observation that conventional sequential multimodal processing frameworks introduce imbalanced and accumulated latency across different modalities, prolonging the overall system delay. In this section, we first introduce the new framework that leverages pipelined sensing and encoding to enable parallel processing of multimodal data streams, and then present the mechanisms to further enhance the performance under dynamic data and system conditions, including an \emph{adaptive multimodal configuration module} and a \emph{cross-modal speculative skipping strategy}.

\subsection{Pipelined Sensing and Encoding Framework}

In the proposed pipelined sensing and encoding framework, we decompose the unimodal encoding process into a sequence of fine-grained \emph{processing units}, each responsible for encoding a localized segment of the input data. Such design reduces end-to-end latency by allowing encoding to proceed incrementally. The encoded features from these units are then aggregated across time to form a unified representation. We first introduce the decomposition framework, and then describe how to preserve accuracy performance through a lightweight and efficient temporal aggregation strategy.

\subsubsection{Decomposition of Encoding Process} 

Traditional multimodal systems typically perform sensing and inference in a sequential manner, where model execution is blocked until all sensor data within a predefined time window are fully acquired. This sequential framework introduces two major limitations: (1) it leads to imbalanced and accumulated delays across modalities due to heterogeneous sensing and processing times, and (2) it increases memory usage due to the need to buffer the entire temporal window before initiating processing.

To address these limitations, \name introduces a fine-grained pipelined decomposition of the unimodal encoding process. As shown in Figure~\ref{fig:system overview}, each pipelined unit is naturally aligned with its corresponding sensing interval, allowing the smallest available data segment (e.g., a single video frame or audio chunk) to be processed immediately upon acquisition. This can be viewed as changing from a full-window processing $y = F(\sum x)$ to a pipelined form $y = \sum f(x)$, where $F(\cdot)$ is a large encoder applied to the entire input sequence, and $f(\cdot)$ is a lightweight encoder operating on individual sensing units.
Moreover, rather than relying on a large model to encode the full temporal window, MMEdge employs lightweight encoders that operate on individual units. For instance, in video tasks, 2D CNN are used to extract spatial features frame-by-frame, replacing more computationally intensive 3D convolutions. Similarly, smaller models are used for time-series data such as audio to reduce latency and resource consumption.
Once all pipelined units within a time window are processed, their encoded features are aggregated temporally and passed to the multimodal fusion and prediction layers. This enables the system to model temporal dependencies effectively while maintaining low latency. For example, in video-based tasks, spatial features are extracted per frame and then aggregated to capture motion dynamics across time.

This decomposition strategy enables parallel execution of data collection and feature encoding across adjacent units, effectively overlapping model inference with the sensing process. As a result, it utilizes idle time between sampling intervals to significantly reduce overall system latency.

\subsubsection{Efficient Temporal Aggregation} 

\begin{figure}
    \centering
    \setlength{\abovecaptionskip}{0.cm}
    \setlength{\belowcaptionskip}{0.cm}
    \includegraphics[width=1\linewidth]{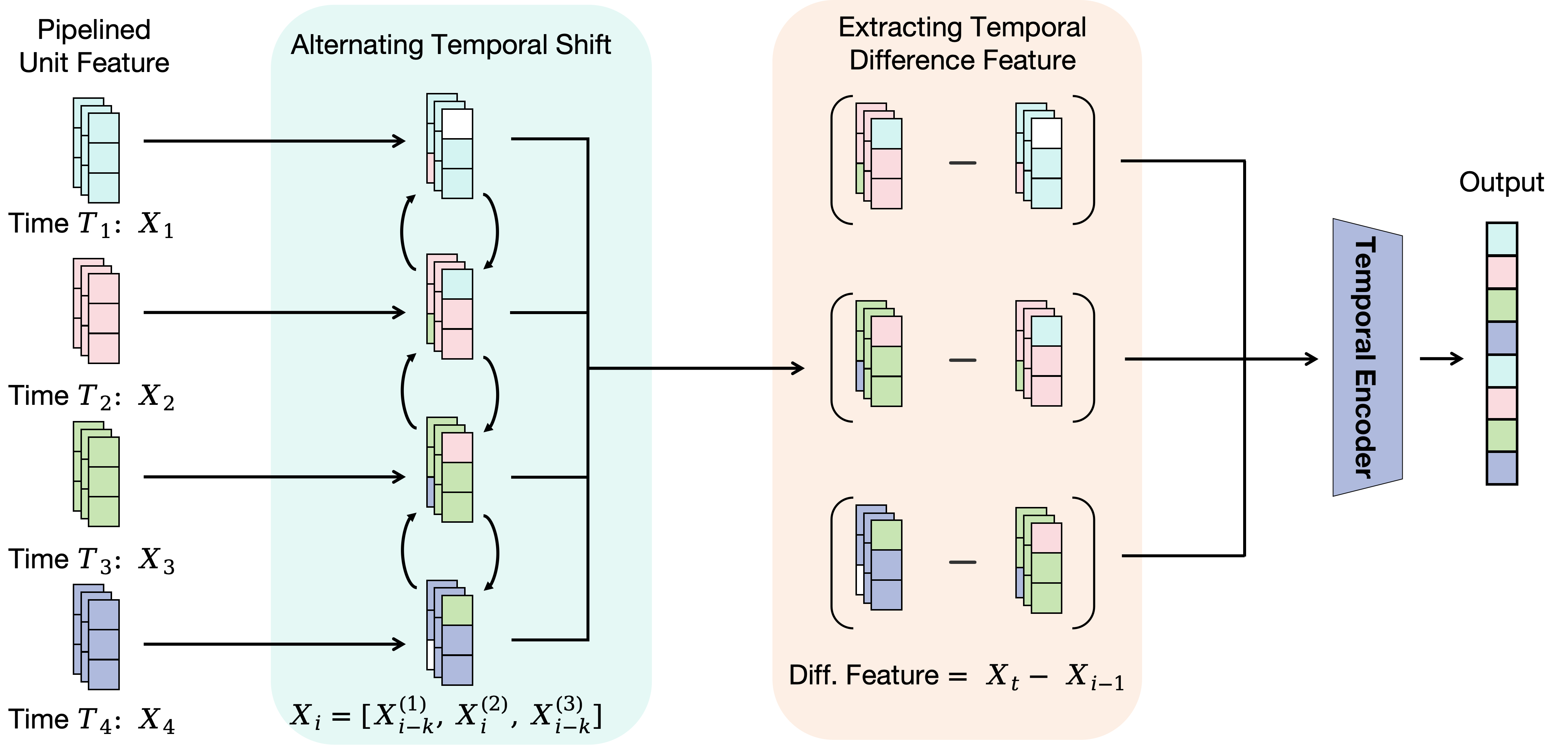}
    \caption{Efficient temporal aggregation through alternating shift across features from neighbor units and extract difference of features to enhance temporal correlation. 
    }
    \label{fig:accuracy drop}
    \vspace{-2em}
\end{figure}

As discussed in Section~\ref{sec:motivation_pipeline}, although the pipelined sensing and encoding framework effectively reduces latency, it may lead to a drop in accuracy due to the reduced granularity of temporal information extraction. To address this issue , we introduce a lightweight temporal aggregation module that captures rich temporal dynamics across different granularities while maintaining low computational overhead.

\noindent
\textbf{Alternating Temporal Shift}. 
In traditional sequential inference pipelines, feature encoders have access to the entire input sequence, allowing them to capture global temporal dependencies during feature extraction. In contrast, our pipelined design processes each data unit (e.g., a video frame or audio chunk) independently upon arrival. While this approach reduces latency, it will also limits the temporal context available to each unit.
To address this limitation, we introduce an alternating temporal shift module that enables each unit to incorporate contextual information from its past and future neighbors. Inspired by temporal shift mechanisms \cite{lin2019tsm}, our method alternates feature channels along the temporal dimension, facilitating lightweight context propagation across adjacent units. 

Specifically, given the feature output $X_i \in \mathbb{R}^{C}$ from pipeline unit $i$, we divide the feature channels into $n$ groups. Here, each ``group'' refers to a contiguous subset of feature channels. For example, when $n=3$, the feature $X_i$ are divided into $X_i = \left[ X_i^{(1)}, X_i^{(2)}, X_i^{(3)}\right]$. 
We then shift the feature groups by replacing them with those from neighboring units, thereby extending the temporal receptive field of each unit through adjacent context integration without increasing model complexity. For instance, the first group is replaced with features from the preceding unit $i-k$, while the third group is substituted with features from the succeeding unit $i+k$.
\begin{align}
    X_i = \left[ X_{i-k}^{(1)}, X_i^{(2)}, X_{i+k}^{(3)} \right],
\end{align}
$X_{i-k}^{(1)}$ and $X_{i+k}^{(3)}$ represent the corresponding channel groups from neighboring units. For the boundary cases (i.e., the first or last unit), they will retain original values as neighbors are unavailable. This alternating temporal shift enables every unit to incorporate contextual information from both earlier and later time steps, effectively approximating the receptive field of full-sequence inference.  


\noindent
\textbf{Extracting Temporal Difference Features}. 
To compensate for the loss of temporal dynamics caused by decomposition, we further enhance temporal modeling by extracting multi-scale difference features across adjacent units. Specifically, we compute temporal differences for features of different units, e.g., $X_t - X_{t-1}$ and $X_t - X_{t-2}$, which approximate short-term and longer-term changes in the feature space. These difference features are then passed through a temporal encoder with global pooling to distill salient temporal variations, enhancing the model’s temporal sensitivity.

By combining alternating temporal shift and temporal difference feature extraction, the model effectively captures temporal dynamics at multiple granularities. Moreover, since both strategies are based on feature augmentation, they enhance temporal coherence while maintaining low computational overhead and introducing no additional latency.

\subsection{Adaptive Multimodal Configuration}

\begin{figure}
    \centering
    \setlength{\abovecaptionskip}{0.cm}
    \setlength{\belowcaptionskip}{0.cm}
    \includegraphics[width=1.0\linewidth]{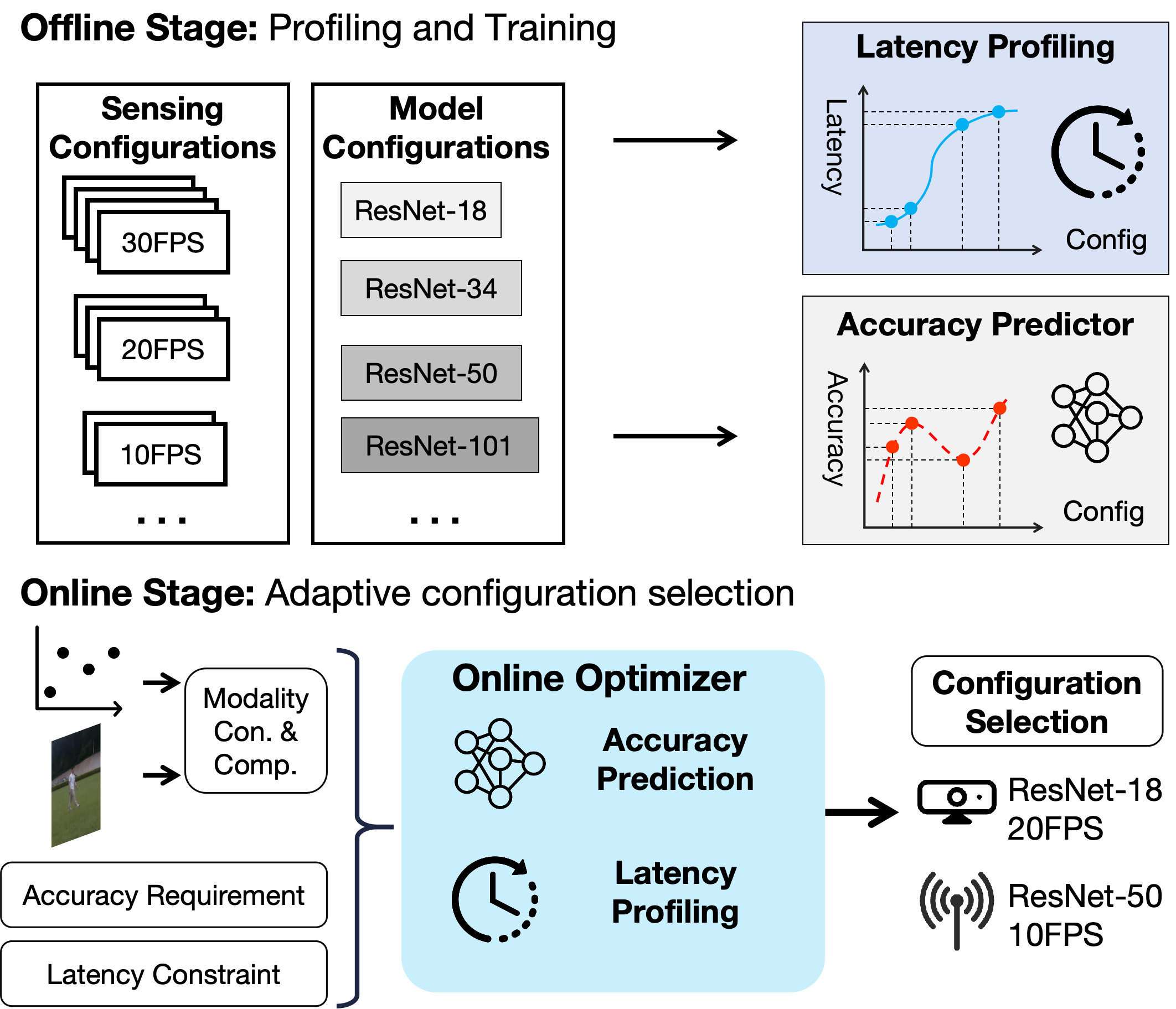}
    \caption{The adaptive multimodal configuration module. (1) Offline stage: latency profiling and train the accuracy predictor. (2) Online stage: the optimizer select configurations based on resource availability and input data.
    }
    \label{fig:adaptive multimodal configurations}
    \vspace{-2em}
\end{figure}

\subsubsection{Motivation and Overview} Although the pipelined sensing and encoding framework reduces end-to-end latency by decomposing the encoding process into a sequence of fine-grained units, the latency of each unit is influenced by both the sensing interval and the encoding delay.
As illustrated in Figure~\ref{fig:pipelined inference}, such design tightly couples sensing and encoding process, with encoding operations occurring during the sensing interval. As a result, the latency of each unit—as well as the overall end-to-end latency—is determined by both the sensing granularity and the model complexity.
For instance, when the sensing granularity is coarse (e.g., low sampling rate) and the feature encoder is lightweight, the sensing interval is relatively long, allowing encoding to complete within the interval. In contrast, with finer data granularity or a more complex encoder, the encoding may exceed the sensing interval, resulting in longer unit delays and accumulated latency across the pipeline.
Therefore, careful and joint selection of both sensing and model configurations is essential to meet latency constraints. This dependency becomes even more critical in multi-modal systems, where feature fusion must wait until all modality-specific encodings are complete.
Moreover, in real-world applications, system dynamics (e.g., CPU contention, thermal throttling) and data dynamics (e.g., varying sample difficulty) can further impact both latency and accuracy. These factors highlight the need for an optimization strategy that dynamically adapts sensing and model configurations across all modalities based on runtime conditions. 


To address these challenges, we design a lightweight optimizer that dynamically selects the optimal combination of sensor and model configurations across different modalities to meet latency constraints while maximizing accuracy.
Figure~\ref{fig:adaptive multimodal configurations} illustrates the architecture of the optimizer, which operates in two stages. In the offline stage,  \name performs profiling of sensing and inference latency across all candidate configurations, and trains an accuracy predictor that estimates the expected accuracy of each configuration given various data samples. 
The profiling is conducted under full execution of the end-to-end multimodal system, including sensing, encoding, and fusion, to capture realistic system dynamics such as CPU scheduling and thermal throttling. This ensures that the profiled latency reflects real-world runtime behaviors rather than idealized isolated measurements.
In the online stage, the optimizer employs an efficient greedy search algorithm to select the optimal configuration based on real-time data inputs and system conditions, leveraging the results given by the latency profiles and accuracy predictor. The optimization is performed under strict latency constraints and incurs negligible runtime overhead, making it suitable for deployment on edge platforms.

We now present the problem formulation of the optimizer and describe the approach for solving it.


\subsubsection{Problem Formulation}

Suppose that we have a sequence of data samples $ X = \{x_1, x_2, x_3, \dots\} $, where each sample consists of data from sensor modalities $ \mathcal{M} = \{ m_1, m_2, m_3, \dots\}$ and is segmented into $N$ units. Each unit comprises two main components: data collection and feature encoding. Therefore, the latency of modality $m_i$ for sample $x_p$ can be calculated as follows:
\begin{align}
    L(x_p^{m_i}) = \max[L_E(x_{p}^{m_i}), L_S] \times N + L_{A}(x_p^{m_i}), i = 1, 2, ..., |\mathcal{M}|,
\end{align}

\noindent where $L_S=\frac{1}{N}$ represents the sensing interval, $L_E(x_p^{m_i})$ denotes the latency of unimodal encoding, $ L_{A}(x_p^{m_i}) $ is the temporal aggregation latency, and $N$ denotes the number of units. Once the unimodal features of different modalities are encoded, they are passed to the multimodal fusion module. Therefore, the end-to-end latency for the multimodal data sample $x_p$ can be defined as:
\begin{align}
    L(x_p) = \max_{m_i \in \mathcal{M}}[L(x_p^{m_i})]  + L_F(x_p),
\end{align}
\noindent where $ L_F(x_p) $ is the latency of multimodal fusion and prediction for sample $x_p$. The objective of the optimizer is to maximize accuracy performance while satisfying latency constraints. Therefore, the optimization problem can be formulated as:
\begin{align}
    \max_{d_{ijk}} \quad & \sum_{i=1}^{|\mathcal{M}|} \sum_{j=1}^{|c_s|} \sum_{k=1}^{|c_m|} \hat{\mathcal{A}}(x^{m_i}, c_s,c_m) \cdot d_{ijk} \label{equ:objective}\\
    \text{s.t.} \quad & L(x) \le T_{\max}, \\
    & \sum_{j=1}^{|c_s|} \sum_{k=1}^{|c_m|} d_{ijk} = 1, \quad \forall i = 1, 2, ..., |\mathcal{M}|. \label{equ:constraint}
\end{align}
\noindent Here $T_{\text{max}}$ denots the latency constraint given by the task requirements. Each candidate configuration consists of a sensing configuration $c_s$ (which controls data granularity, e.g., frame rate for video or chunk size for audio) and a model configuration $c_m$ (which controls model complexity, e.g., model size of feature encoder). We define a binary decision variable $d_{ijk} \in \{0,1\}, \forall i = 1, 2, ..., |\mathcal{M}|$ , where $\mathcal{M}$ is the set of modalities.  The variable $d_{ijk}=1$ indicates that the $j$-th sensing configuration and $k$-th model configuration are selected for modality $m_i$. 
The accuracy predictor $\hat{\mathcal{A}}(\cdot)$ is trained offline using labeled samples to learn the relationship between accuracy performance and data characteristics, given the selected sensing and model configurations. Unlike latency, which mainly depends on device-specific factors and can be profiled and stored in a lookup table, accuracy is highly data-dependent and varies with unseen data samples. Therefore, the predictor will be able to estimate accuracy even for unseen inputs during online optimization.

\subsubsection{Lightweight Accuracy Predictor.} In the objective formulation of the problem (i.e., Equation~\ref{equ:objective}), $A(\cdot)$ denotes an accuracy predictor that estimates the expected performance of different configurations on varying data inputs. However, directly performing this estimation from raw multimodal sensor data $x$ would require a large and computationally expensive model, making it impractical for real-time optimization in edge deployment. To address this, we propose a lightweight accuracy predictor that operates on informative features extracted from the raw multimodal data. Specifically, we introduce two auxiliary metrics designed to capture consistent and complementary information across modalities, which are important to influence the effectiveness of multimodal learning systems \cite{ouyang2022cosmo, wu2024adaflow}. These metrics enable the accuracy predictor to make accurate performance estimates without incurring significant computational overhead. Specifically, we define the two types of multimodal information as follows:

\begin{itemize}
    \item \textbf{Modality Consistency $ \text{Cons}(x_i) $}, which measures the alignment between modalities by computing the average pairwise cosine similarity among their unimodal features:
    \begin{align}
        \text{Cons}(x_i) = \frac{\mathbf{f}^{(m_1)} \cdot \mathbf{f}^{(m_2)}}{\|\mathbf{f}^{(m_1)}\| \cdot \|\mathbf{f}^{(m_2)}\|},
    \end{align}
    here $\mathbf{f}^{(m_i)}$ is the feature of the first frame from modality $m_i$.
\item  \textbf{Modality Complementarity $\text{Comp}(x_i)$}, which is defined as the inverse of consistency. A higher complementarity score indicates greater dissimilarity between modalities, suggesting that they may provide richer and more diverse information when fused. It is denoted as:
    \begin{align}
        \text{Comp}(x_i) = 1 - \text{Cons}(x_i)
    \end{align}
\end{itemize}
These two metrics are incorporated into the accuracy predictor $ \hat{\mathcal{A}}(x_i, c) = f_{\theta}(\text{Cons}(x_i), \text{Comp}(x_i), c) $, which maps the high-dimensional multimodal input into a compact 2D representation, thus enabling efficient and lightweight configuration selection during inference.

\begin{algorithm}
    \caption{Multimodal Configuration Optimizer  
    }
    \label{alg:optimizer}
    \smallskip
    \textbf{Initialize:} Configuration sets $\mathcal{C}_s$, $\mathcal{C}_m$, Accuracy predictor $f_A$, latency lookup table $\mathcal{L}$, latency target $T_{\max}$, modality set $\mathcal{M}$, greedy search algorithm GSA.
    
    \begin{algorithmic}[1]
    \STATE \textbf{function} \textsc{Optimizer}($x_i$):
    \STATE \quad $\text{Cons}(x_i) \leftarrow$ Compute \textbf{modality consistency} cos($x_i$)
    \STATE \quad $\text{Comp}(x_i) \leftarrow $ Compute \textbf{modality complementary} $1 - \text{cos}(x_i)$
    \STATE \quad $\hat{A} \leftarrow$ \textbf{Accuracy prediction} $f_A(\text{Cons}, \text{Comp}, c)$
    \STATE \quad $\hat{L} \leftarrow$ Latency Prediction $\mathcal{L}(c)$,
    \STATE \quad $c_i^* \leftarrow $ Find optimal configuration by GSA($\hat{A}$, $\hat{L}$, $T_{max}$)
    \STATE \quad \textbf{return} $c_i^*$
    
    \vspace{0.5em}
    \
    \ \textit{\# Online execution for streaming input}
    \WHILE{system is running}
    \STATE \quad Collect data $x_i$ by sensors
    \STATE \quad $c_i^* \leftarrow$ \textsc{Optimizer}($x_i$)
    \STATE \quad Apply configuration $c_i^*$ for sensing and inference
    \ENDWHILE
    \end{algorithmic}
\end{algorithm}

\subsubsection{Online Multimodal Configuration optimizer}
To solve the optimization problem defined in Equation~\ref{equ:objective}–\ref{equ:constraint}, we employ a greedy search algorithm to identify the optimal configuration $c^*$ under a given latency constraint $T_{\text{max}}$. The goal is to maximize the estimated accuracy $\hat{\mathcal{A}}(x_i, c)$ as predicted by the lightweight accuracy predictor, while ensuring that the total end-to-end latency remains within the specified constraint $T_{\text{max}}$. Algorithm~\ref{alg:optimizer} outlines the optimization process.

To enable efficient runtime decision-making, we pre-profile the end-to-end latency $L(c)$ offline for all candidate configurations on the target platform, and store the results in a lookup table. During online inference, the optimizer performs a greedy search guided by the outputs of the accuracy predictor and the latency lookup table. This design ensures that the total runtime overhead of online optimizer remains minimal, typically within a few milliseconds. By integrating the lightweight accuracy predictor that adapts to varying data inputs and a latency profiler that accounts for system dynamics, our solution enables efficient and fine-grained configuration selection for real-time multimodal inference.

\subsection{Cross-Modal Speculative Skipping}

Due to varying processing speeds across modalities, multimodal systems often experience modality asynchronization, where faster modalities (e.g., audio) complete sensing and encoding well before slower ones (e.g., video). This can significantly increase end-to-end latency of the multimodal system, as the system must wait for the unimodal encoding of all modalities to complete before performing feature fusion, even when partial inputs from some modalities may already be sufficient for accurate inference.
To address this challenge, we propose a \textit{cross-modal speculative skipping} mechanism, which allows the system to terminate inference early when partial data from slower modalities is likely sufficient for reliable prediction. The key idea is to employ a lightweight gating classifier that dynamically decides whether to wait for additional data from the slower modality or to proceed with inference using the currently available information. This enables the system to reduce latency without compromising prediction accuracy under various system and data dynamics.

\subsubsection{Light-weight Gating Classifier}

The gating classifier is trained offline to predict whether early termination, based on partial features of slower modalities, would yield the same decision as full-input inference.
Instead of relying on static thresholds, the gating classifier learns the decision boundary directly from data, capturing complex cross-modal dependencies between fast and slow modalities. This design allows it to generalize across unseen scenarios and adapt to diverse content conditions, achieving reliable early termination without manual tuning.

The input for the gating classifier consists of: (1) $f_{\text{fast}}$: complete features from the fast modality (e.g., audio), and (2) $f_{\text{slow}}$: partial features from the slow modality (e.g., 50\% or 70\% of video frames).
These features are concatenated as $[f_{\text{fast}}, f_{\text{slow}}]$ and fed into a lightweight classifier to produce a prediction $\hat{y}$. A binary supervision label is then generated to indicate whether the prediction from partial inputs matches the prediction from full inputs, guiding the gating classifier to learn when early inference is safe.
For the gating classifier, we employ a two-layer multi-layer perceptron (MLP) with dropout regularization and a sigmoid activation function at the output layer. The model is trained using the binary cross-entropy loss, defined as:
\begin{align}
    \mathcal{L} = - y \log(\hat{y}) + (1 - y) \log(1 - \hat{y}),
\end{align}
where  $y$ is the ground truth label indicating whether early termination yields the same prediction as full inference ($y=1$) or not ($y=0$), and $\hat{y}$ is the gating classifier's predicted probability.
To effectively train the gating classifier, we construct a diverse training set by exploring a range of system configurations that vary in both sensing granularity and model complexity. This diversity helps the classifier generalize across different multimodal scenarios and make robust early skipping decisions.

\begin{figure}
    \centering
    \setlength{\abovecaptionskip}{0.cm}
    \setlength{\belowcaptionskip}{0.cm}
    \includegraphics[width=1\linewidth]{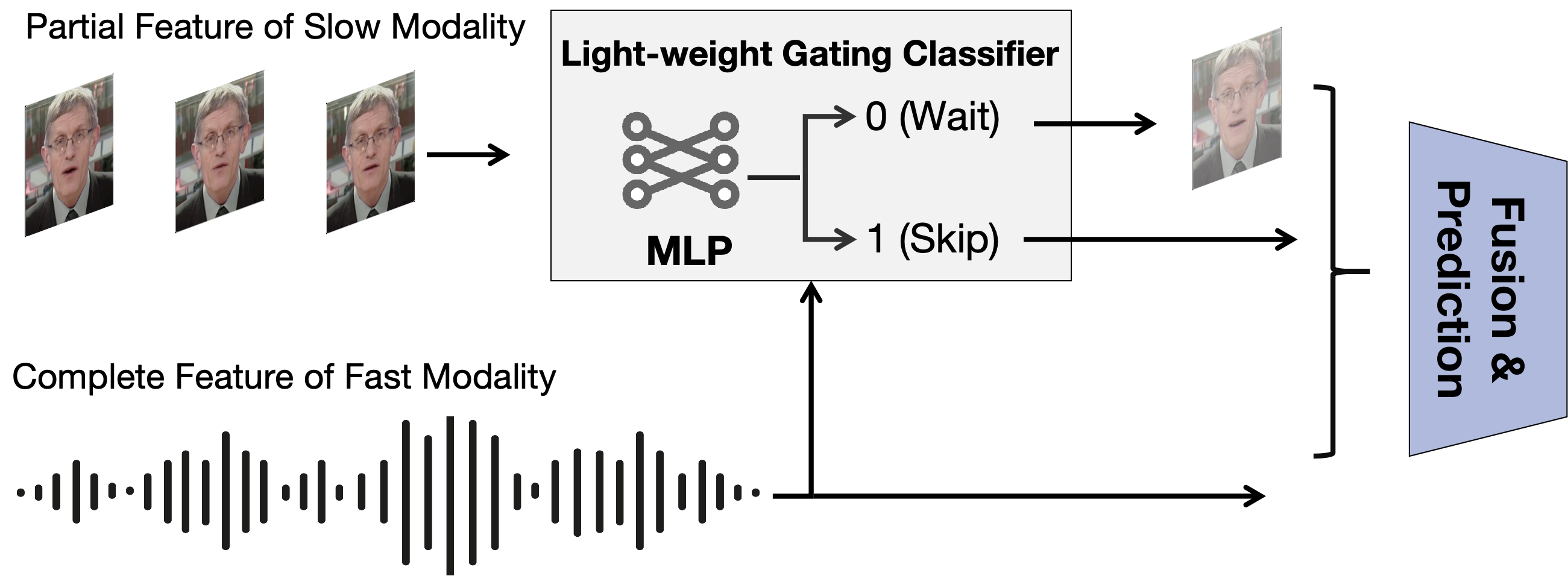}
    \caption{Cross-modal speculative skipping: leveraging complete features from fast modalities and partial results from slower pipelines to trigger early inference termination, reducing end-to-end latency under modality asynchrony.  
    }
    \label{fig:early exit}
    \vspace{-2em}
\end{figure}

\subsubsection{Online Modality Skipping}
At runtime, the encoders for each modality operate as independent pipelines. Features from faster modalities (e.g., audio) become available earlier and are stored in a shared buffer. Once the encoding process for the slower modality (e.g., video) reaches a predefined checkpoint, such as after processing 50\% or 70\% of all input units, the system evaluates the confidence of early prediction using the trained gating classifier, based on the combined features $[f_{\text{fast}}, f_{\text{slow}}]$.
The classifier outputs a probability score $p \in [0,1]$, representing the likelihood that early termination will yield a correct prediction. If $p > \tau$ (e.g., $\tau = 0.5$), the system triggers early stopping, skipping the remaining units of the slower modality to reduce computation and latency. The threshold $\tau$ controls the confidence level required for early termination and can be tuned to balance latency and accuracy. In our implementation, we adopt $\tau = 0.5$, following the standard decision boundary used in binary classification. If the confidence score does not exceed the threshold, the system continues processing until all modality inputs are fully encoded.

This design is also naturally integrated with the pipelined sensing and encoding framework, as it enables joint skipping of both redundant data collection and unnecessary processing. By leveraging complete information from faster modalities, the system avoids gathering and encoding data from slower modalities when it is unlikely to improve prediction accuracy, thereby reducing computational latency. 
For example, in an audio-visual speech recognition task, if the audio features alone yield a confident prediction, the system can skip the remaining video frames to reduce latency without compromising performance.
Moreover, the gating classifier is highly efficient as it operates solely on extracted features and performs a simple binary classification, introducing negligible runtime overhead (less than 2 ms in our implementation). 


\section{Experiments}
\label{sec:evaluation}


\subsection{Testbed and Datasets}

\begin{figure}
    \centering
    \setlength{\abovecaptionskip}{0.cm}
    \setlength{\belowcaptionskip}{0.cm}
        \begin{subfigure}{0.5\linewidth}
            \setlength{\abovecaptionskip}{0.cm}
    \setlength{\belowcaptionskip}{0.cm}
            \includegraphics[width=1\linewidth]{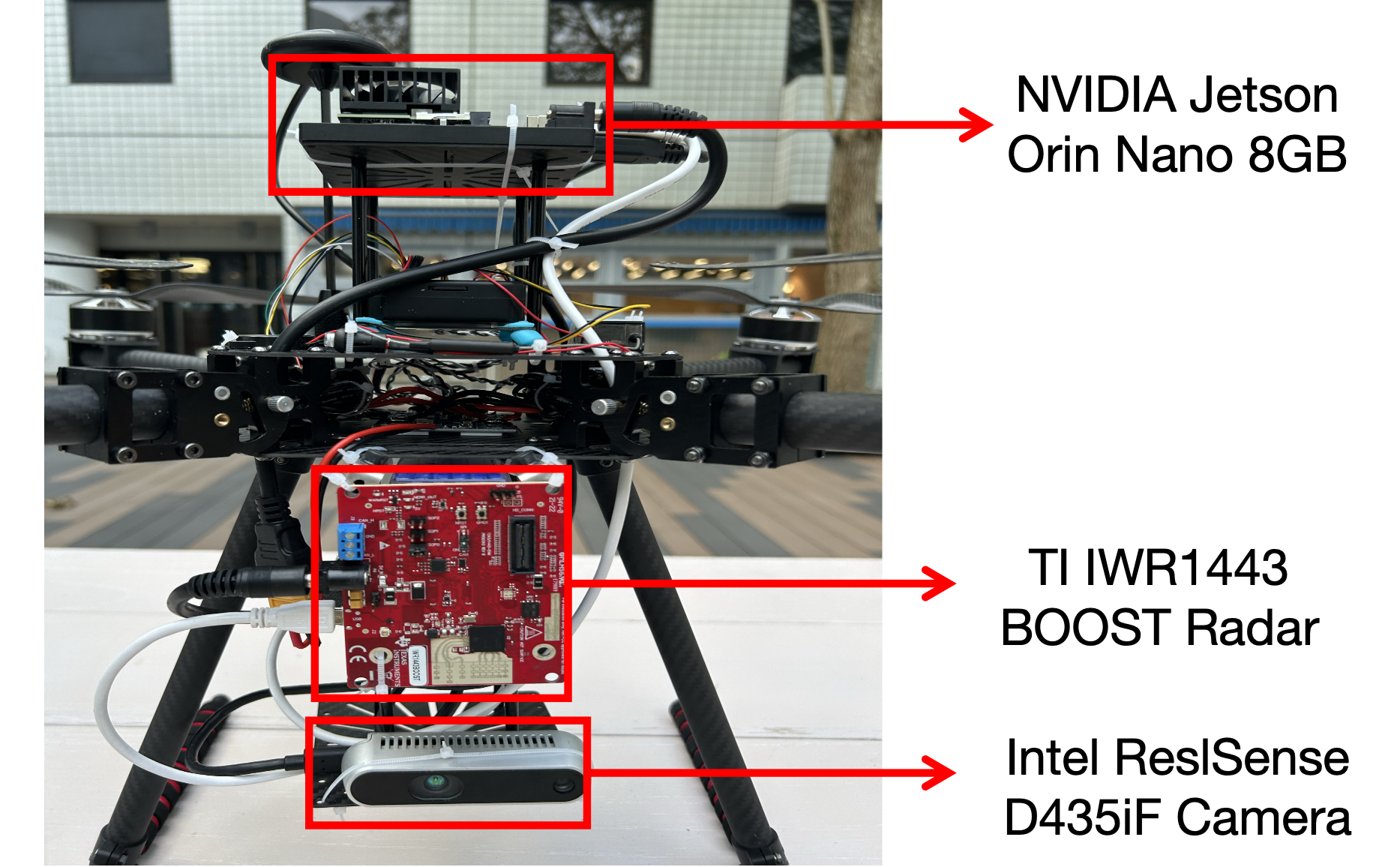}
            \caption{Multi-modal UAV Testbed.}
            \label{fig:multimodal uav testbed}
        \end{subfigure}%
        \begin{subfigure}{0.5\linewidth}
            \setlength{\abovecaptionskip}{0.cm}
    \setlength{\belowcaptionskip}{0.cm}
            \includegraphics[width=0.8\linewidth]{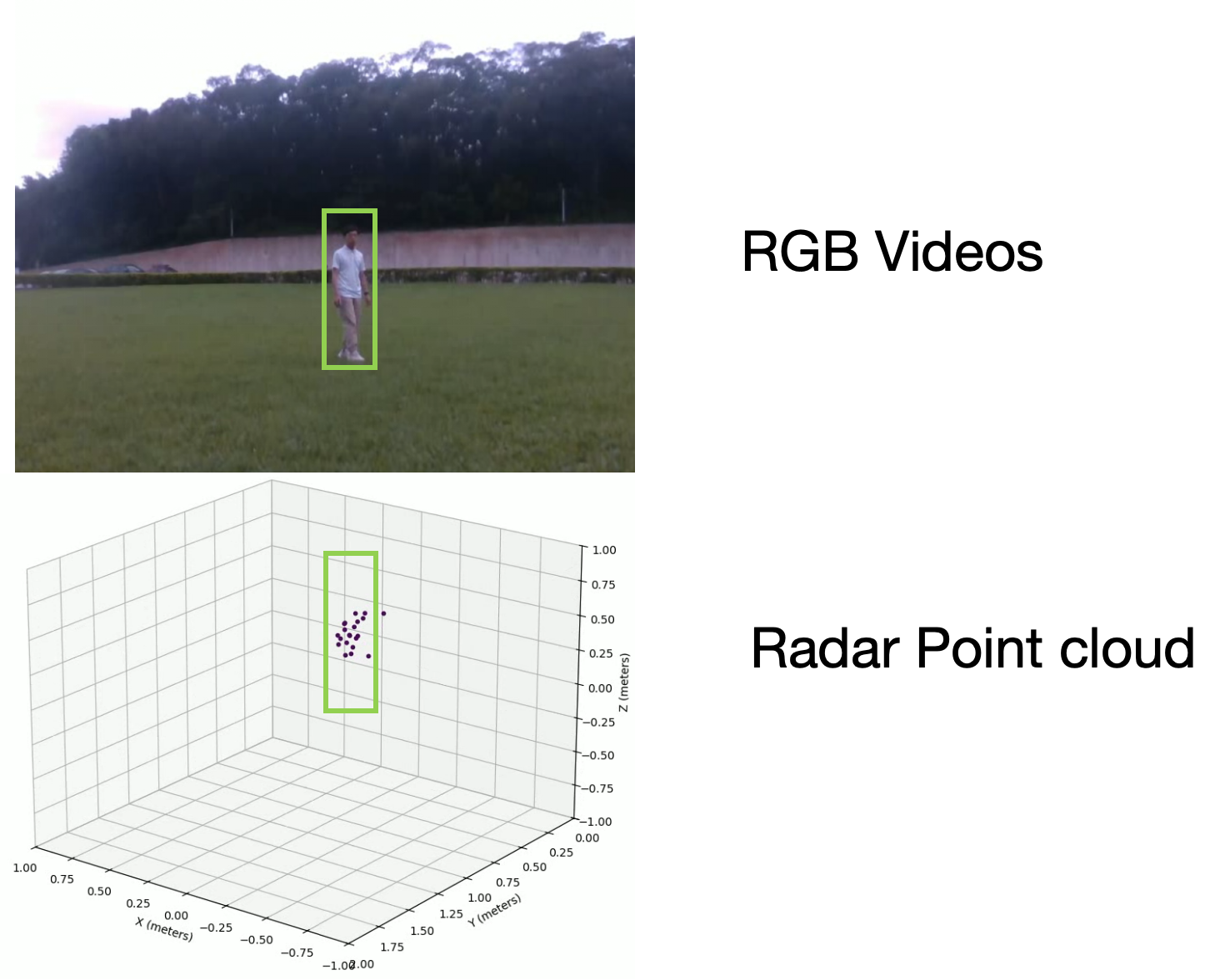}
            \caption{Collected data.}
            \label{fig:example of collected data}
        \end{subfigure}
    \caption{The real-world UAV testbed for human tracking.
    }
    \label{fig:uav testbed}
    \vspace{-1em}
\end{figure}

\begin{table}
    \centering
    \resizebox{\linewidth}{!}{
    \begin{tabular}{lcccc}
        \toprule
        \textbf{Dataset} & \textbf{Task} & \textbf{Modalities} & \# \textbf{Classes} & \# \textbf{Samples} \\
        \midrule
        LRW & SR  & RGB+A        & 50 & 50,000 \\
        \begin{tabular}[c]{@{}l@{}}NuScenes-QA-Mini\end{tabular} & VQA & RGB+L+T & 30 & 4,458 \\
        Self-collected & HT  & RGB+R        & 2  & 5,687 \\
        \bottomrule
    \end{tabular}
    }
    \caption{Summary of the datasets. SR: speech recognition. VQA: visual question answering. HT: human tracking. A: audio. L: Lidar. T: text. R: radar.
    }
    \label{tab:datasets}
    \vspace{-3em}
\end{table}

\noindent \textbf{Real-world UAV Testbed}. 
We built a UAV-based multimodal testbed and deployed MMEdge on it to evaluate end-to-end performance under resource constraints in real-world, time-critical applications. As shown in Figure~\ref{fig:uav testbed}, the testbed integrates an Intel RealSense D435iF camera \cite{IntelRealSenseD435i} and a TI IWR1443BOOST mmWave radar \cite{TI_IWR1443_2018} for multimodal data collection, both connected to an onboard NVIDIA Jetson Orin Nano edge computer~\cite{jetson_orin}. The task is to track human subjects on the ground using camera and radar data captured by the UAV in flight\footnote{All the data collection was approved by the Institutional Review Board (IRB) of the authors’ institution}.

Unlike most public datasets that only provide recorded data samples, our testbed supports continuous, end-to-end data collection and model inference on devices, enabling the capture of temporal dynamics and runtime variability inherent in real-world deployments. The system operates throughout UAV flight, where real-world challenges—such as unstable voltage and in-flight vibrations—can trigger thermal throttling, disrupt sensor sampling, and degrade data quality. These conditions introduce variable system load and unpredictable latency, complicating the delivery of reliable, real-time multimodal inference.
Such \emph{system dynamics} are essential for evaluating the adaptability of MMEdge to fluctuations in sensing quality, resource availability, and task complexity.
To evaluate MMEdge's performance under real-world \emph{data dynamics}, we design multiple test scenarios with variations in environment, lighting conditions, and distances to human subjects. In each scenario, the UAV collects multimodal data continuously for approximately 5 minutes, with the RGB camera operating at 30 FPS and the radar at 20 Hz. After data preprocessing and filtering, we obtain 14,600 valid frames, which are segmented into 5,687 samples using a 0.5-second sliding window with a 0.1-second stride. We record multimodal data (RGB videos, radar point clouds) and synchronized systems logs (e.g., voltage, CPU/GPU usage, flight states) during UAV flights. To ensure fair and repeatable evaluations, we implement the MMEdge system and baselines approaches on the UAV testbed using collected multimodal data and configure the Nvidia edge computer based on recorded system logs, enabling end-to-end performance assessment under dynamic resource-constrained conditions.

\noindent \textbf{Public datasets}.
We also evaluate MMEdge on two public multimodal datasets: Lip Reading in the Wild (LRW)~\cite{chung2017lip} and NuScenes-Mini-QA~\cite{qian2024nuscenes}.
LRW is a large-scale audio-visual dataset for speech recognition based on audio and video data, which includes over 500 spoken word classes and contains approximately 55,000 data samples.
NuScenes-Mini-QA is a multimodal question answering (QA) dataset derived from the NuScenes autonomous driving scenes. It features temporal sequences of RGB images and LiDAR point clouds, as well as corresponding QA pairs, enabling question answering tasks related to dynamic driving scenarios.


\vspace{-1em}
\subsection{Experiment Setup}

\noindent \textbf{Devices}. For evaluations on the UAV testbed, we implement MMEdge on NVIDIA Jetson Orin Nano~\cite{jetson_orin}. For evaluations on public datasets, we implement MMEdge on the NVIDIA Jetson Xavier NX~\cite{jetson_xavier}, which features a 6-core ARM v8.2 64-bit CPU, a 384-core Volta GPU with 48 Tensor Cores, and 16 GB of shared LPDDR4x memory. This device is widely used in real-world edge AI applications, such as speech recognition and autonomous driving.
Model training is conducted offline on a high-performance server equipped with dual AMD EPYC 7K62 processors (48 cores, 96 threads) and eight NVIDIA RTX 4090 GPUs, each with 24 GB of memory.

\noindent \textbf{Implementation}. For evaluations on the UAV testbed, we implement MMEdge in an end-to-end manner. For evaluations on public datasets, we follow prior works~\cite{xu2020approxdet, li2021low} to simulate data collection and resource dynamics. Specifically, we run background multimodal data collection processes to emulate the sensing overhead on edge devices.
During evaluations on public datasets, data are loaded while preserving the original frame rate by inserting fixed sleep intervals to simulate sensing delays.
To simulate dynamic resource availability in real-world scenarios, we conduct experiments under various conditions by limiting CPU usage via Linux \texttt{cgroup}.

\noindent \textbf{Baselines}. We compare \name with the below baselines. 
\begin{itemize}
    \item \textbf{Blocking Inference} \cite{wang2023patch}, which waits for all modality inputs before making prediction, ensuring complete data at the cost of higher latency.
    \item \textbf{Non-Blocking Inference} \cite{wu2024adaflow}, which processes available modalities as they arrive, enabling early predictions to reduce latency at the cost of incomplete information.
    \item \textbf{Modality Gaiting} \cite{hou2023smg}, which dynamically selects input modalities before inference to reduce sensing and computation overhead. We implemented it by dynamically reducing the data rate of less informative modalities.
    \item \textbf{Imputation-Based Methods} \cite{li2021low, wu2024adaflow}, which imputes missing data of slower modalities to enhance inference robustness under asynchronous data inputs. We mplemente it by training a GAN  to impute the features of the slow modality.
    \item \textbf{Model Selection}~\cite{rastikerdar2024cactus}, which dynamically chooses model branches based on input data complexity to balance accuracy and efficiency. We implemented it by dynamic model switch.
    \item \textbf{Pipelined Inference Max}, which executes our new pipelined sensing and encoding framework with the largest configuration (e.g., highest data rate and largest models), offering peak accuracy at the cost of maximum latency.
    \item \textbf{Pipelined Inference Min}, which uses the smallest configuration (e.g., lowest data rate and smallest models) in our new pipelined sensing and encoding framework, minimizing latency at the cost of reduced accuracy.
\end{itemize}

\noindent \textbf{Evaluation metrics}. We evaluate the end-to-end system latency and task-specific model accuracy.
The end-to-end latency is defined as the time between the acquisition of the first data unit of the sample ($T_0$) and the completion of its prediction ($T_{\text{end}}$), excluding the duration of the sensing time window ($t_w$), i.e., the latency $ L = ( T_{end} - T_0 ) - t_w $. 
For accuracy performance, we assess top-1 and top-5 accuracy for the speech recognition task on the LRW dataset, and for question answering generation task for the NuScenes-Mini-QA dataset. We calculate Intersection over Union (IoU) for the human tracking task in our real-world UAV testbed.

\noindent \textbf{Sensing and Model Configurations}. Here we introduce the setting of different sensing and model configurations  across datasets.  For RGB, radar, and LiDAR modalities in our datasets, we employ three levels of model complexity using ResNet-18, ResNet-34, and ResNet-50. For audio, we design three CNN-based encoders with varying channel sizes and depths, named as small, medium and large. We use the facebook/OPT-125M \cite{zhang2022opt} model as the LLM for answer generation in the NuScenes-Mini-QA dataset. In our self-collected dataset, we use ResNet for radar and YOLO series models for RGB \cite{redmon2016you}. For sensing configurations, we vary the chunk size for audio (800, 1000, 1200). The RGB frame rates are set to 20, 25, and 29 FPS in LRW; 2, 6, and 12 FPS in NuScenes-Mini-QA; and 10, 20, and 30 FPS in our self-collected dataset. Radar sampling rates in our dataset are 5, 10, and 20 Hz, and LiDAR sampling rates in NuScenes-Mini-QA are 2, 10, and 20 Hz.

\begin{figure}
    \centering
    \setlength{\abovecaptionskip}{0.cm}
    \setlength{\belowcaptionskip}{0.cm}
    \includegraphics[width=1.0\linewidth]{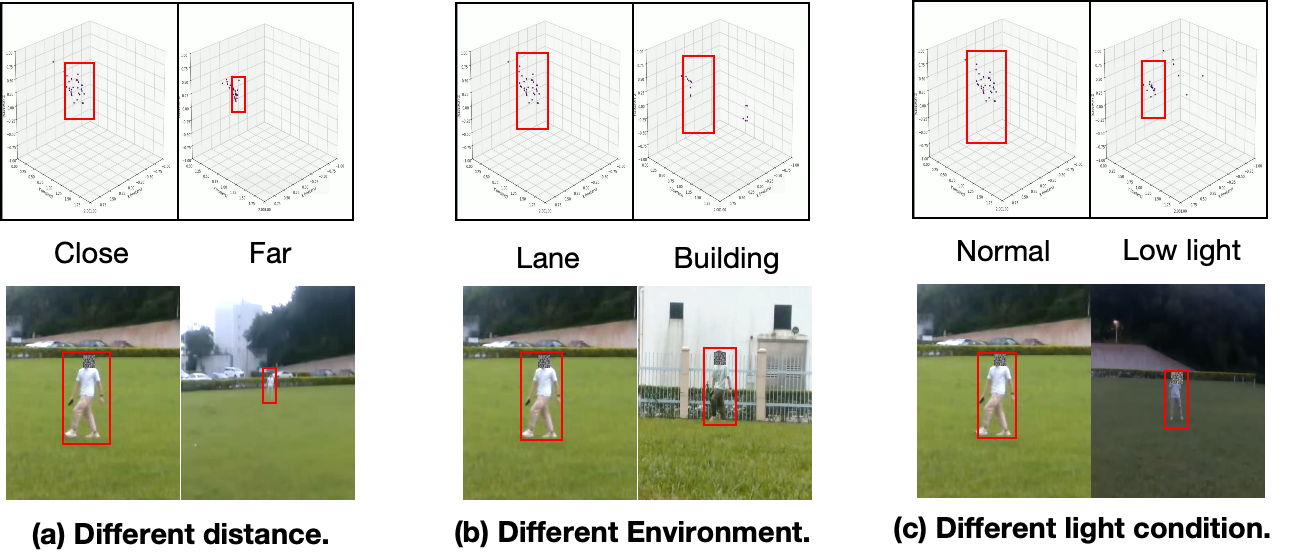}
    \caption{Detection results in different scenarios. 
    }
    \label{fig:sample dynamic}
    \vspace{-5em}
\end{figure}

\begin{figure}
    \centering
    \setlength{\abovecaptionskip}{0.cm}
    \setlength{\belowcaptionskip}{0.cm}
    \includegraphics[width=1.0\linewidth]{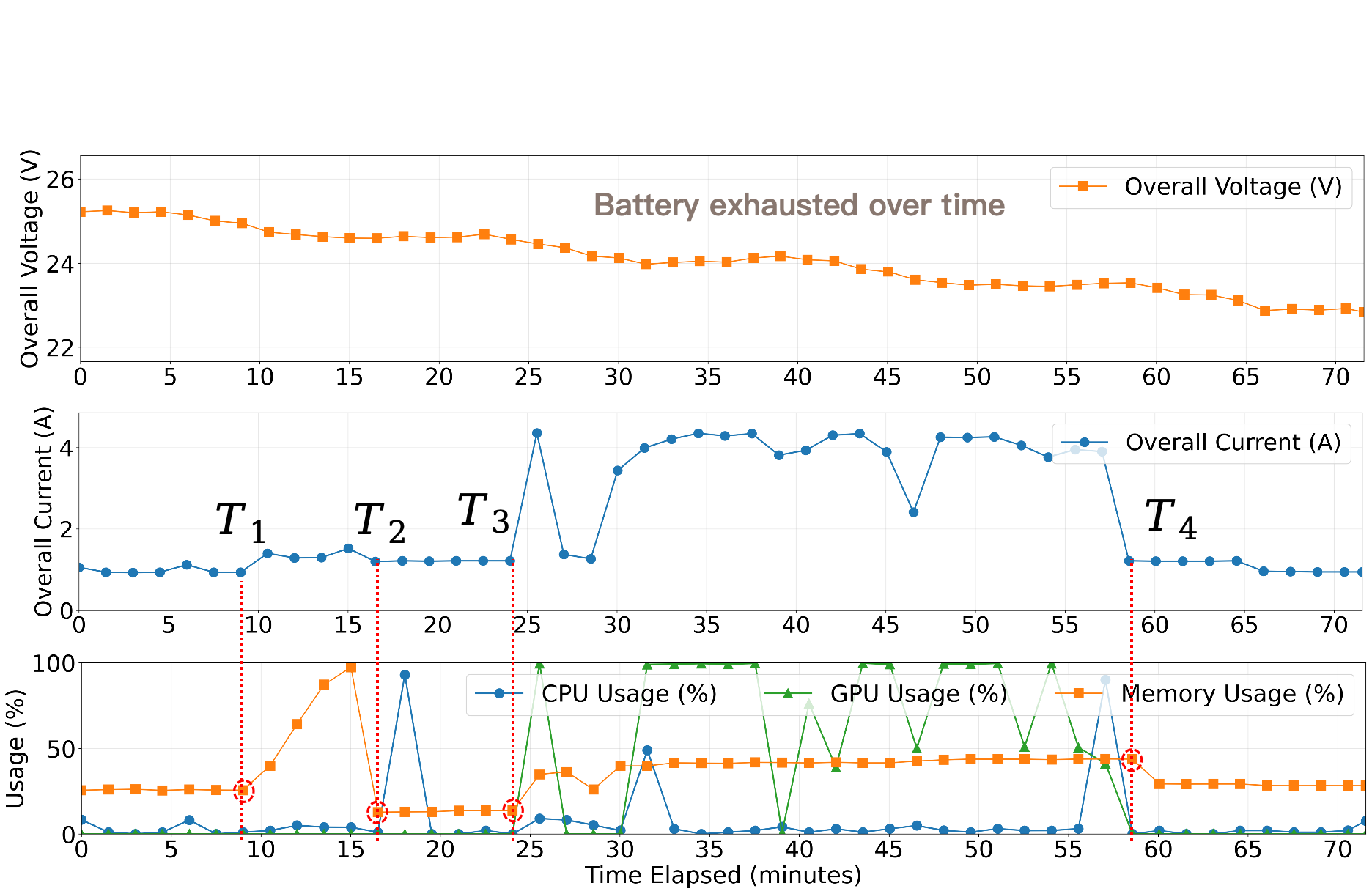}
    \caption{System dynamics on the UAV testbed. $T_1$: data collection starts. $T_2$: the UAV takes off. $T_3$: model inference starts. $T_4$: model inference ends.
    }
    \label{fig:system dynamic}
    \vspace{-4em}
\end{figure}

\begin{figure*}
    \vspace{-2mm}
    \begin{minipage}{1\textwidth}
        \centering
        \includegraphics[width=1\linewidth]{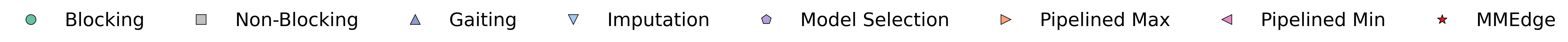}
    \end{minipage}
    \vspace{-3mm}
    
    \centering
    \begin{subfigure}{0.32\linewidth}
        \setlength{\abovecaptionskip}{0.cm}
        \setlength{\belowcaptionskip}{0.cm}
        \includegraphics[width=1\linewidth]{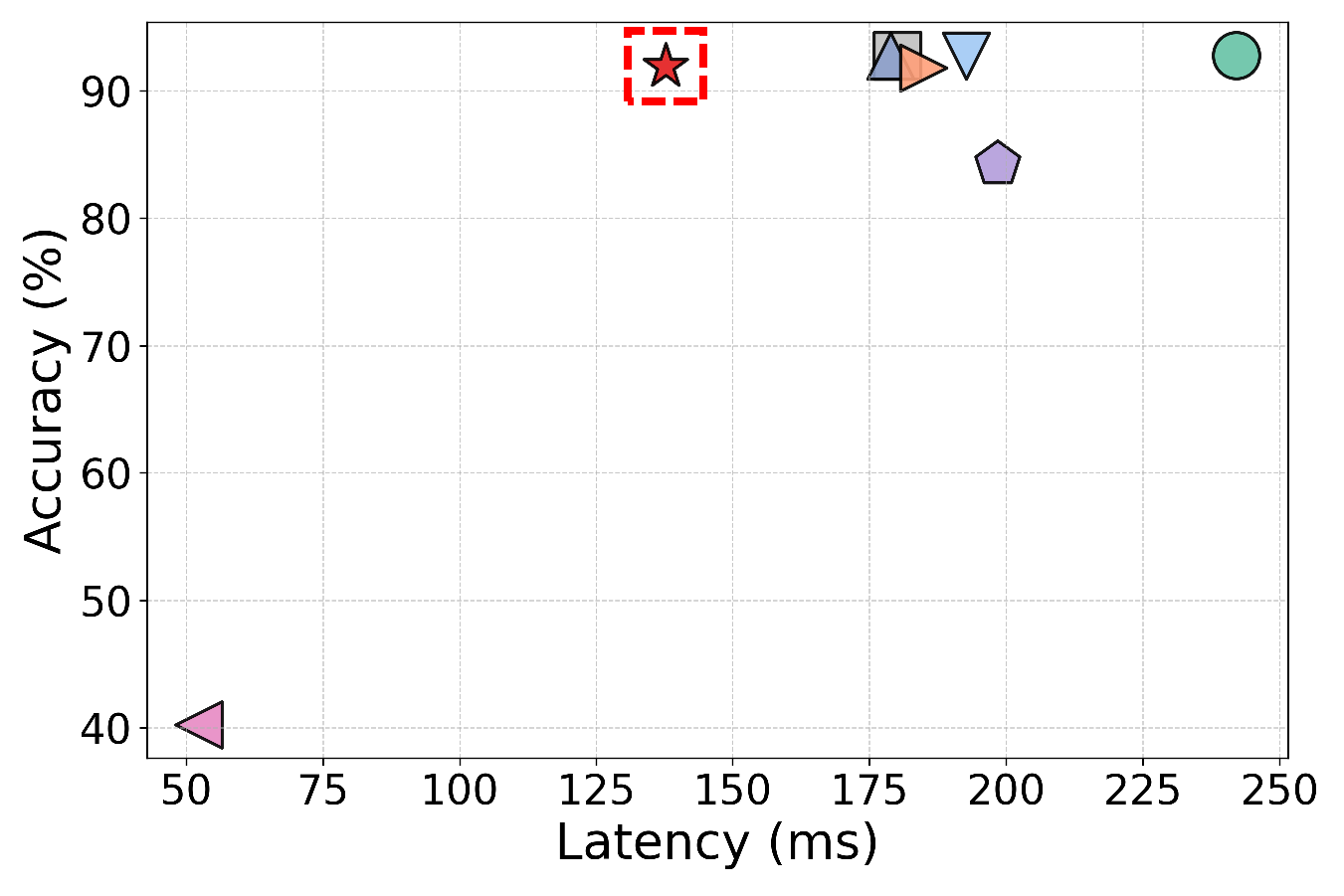}
        \caption{LRW dataset.}
        \label{fig:overall performance lrw}
        \vspace{-3mm}
    \end{subfigure}
    \begin{subfigure}{0.32\linewidth}
       \setlength{\abovecaptionskip}{0.cm}
        \setlength{\belowcaptionskip}{0.cm}
        \includegraphics[width=1\linewidth]{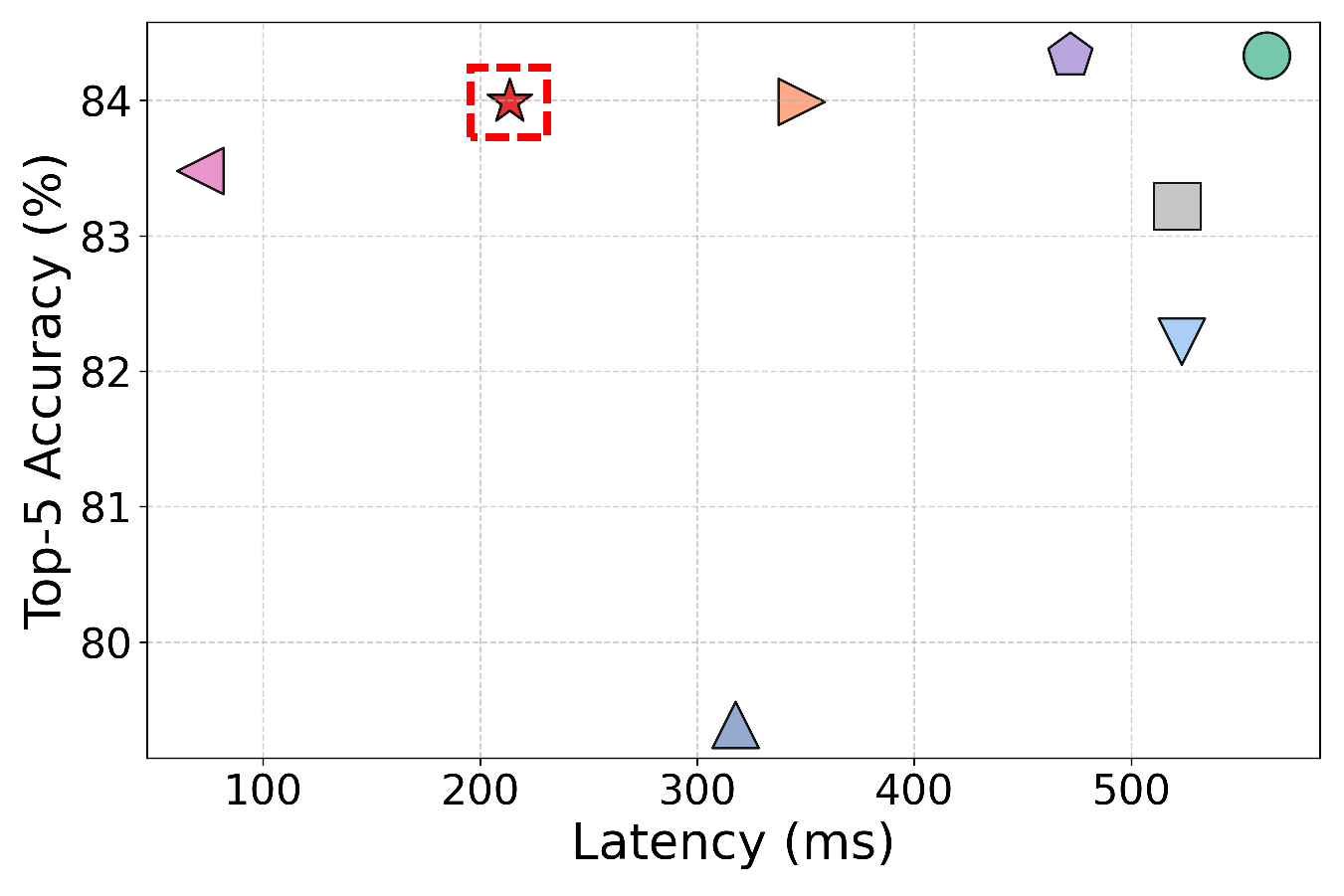}
        \caption{NuScenes-QA-Mini dataset.}
        \label{fig:overall performance nuscenes}
        \vspace{-3mm}
    \end{subfigure}
    \begin{subfigure}{0.32\linewidth}
       \setlength{\abovecaptionskip}{0.cm}
        \setlength{\belowcaptionskip}{0.cm}
        \includegraphics[width=1\linewidth]{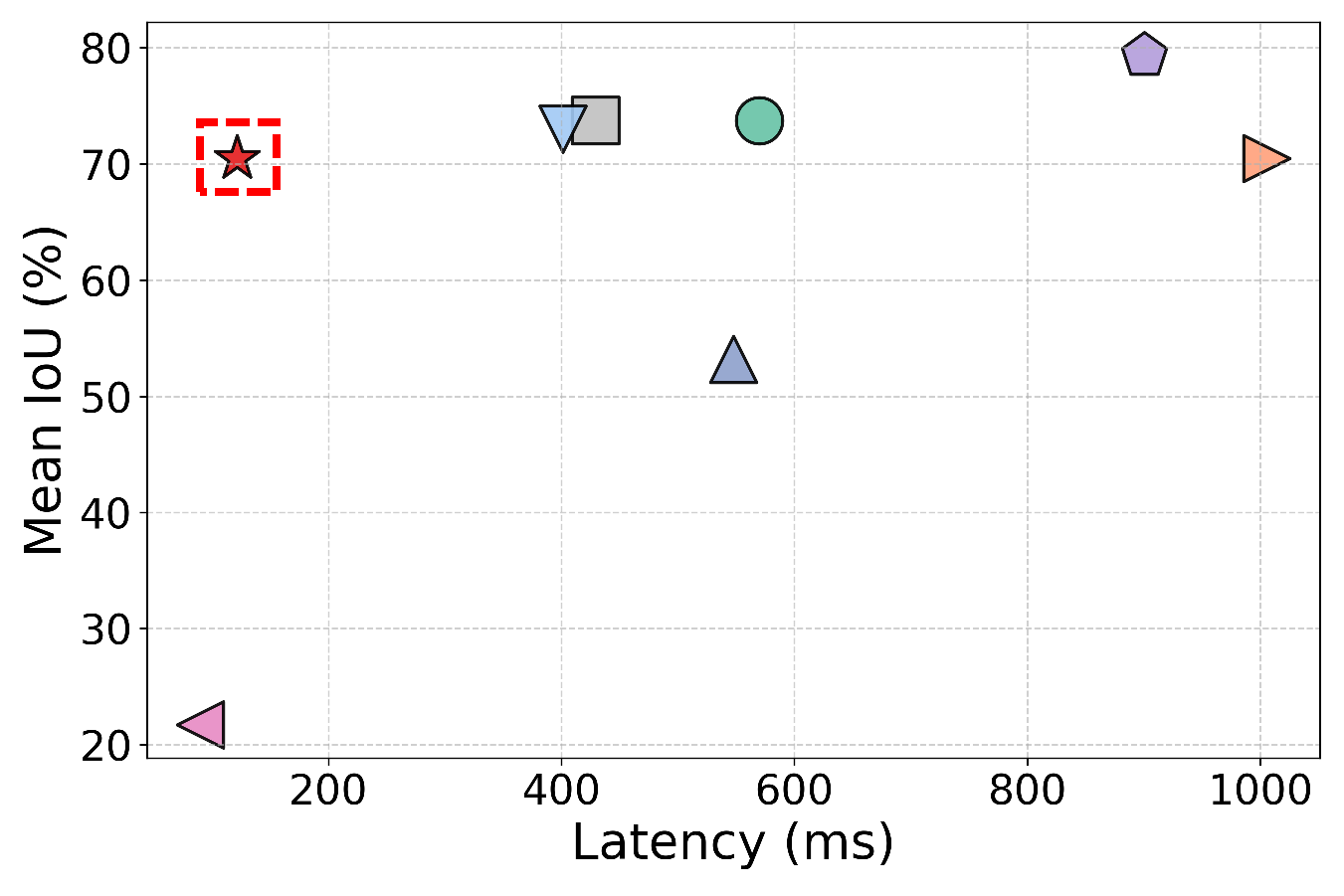}
        \caption{Self-collect dataset.}
        \label{fig:overall performance uav}
        \vspace{-3mm}
    \end{subfigure}
    
    \caption{Overall performance on different datasets. The upper-left region indicates methods that achieve high accuracy with low latency. Compared to baselines, MMEdge consistently achieves an optimal trade-off between these two metrics.}
    \label{fig:overall performance}
    \vspace{-1.5em}
\end{figure*}

\subsection{Case Study on the UAV Testbed}


\textbf{Data dynamics.} To capture variability of data input in real-world applications, we conduct the data collection in multiple test scenarios. 
Figure~\ref{fig:sample dynamic} illustrates the dynamics of data samples across three aspects: the distance between the UAV and the human, the surrounding environment, and lighting conditions. These variations pose significant challenges for multimodal-based human tracking. Specifically, long-range targets result in smaller and sparser point cloud returns from the radar and lower-resolution visual features. Diverse backgrounds (e.g., buildings, vegetation) introduce distractors that can confuse detection models. In low-light conditions, visual modality becomes unreliable, increasing reliance on complementary sensing of radar data. 


\textbf{System dynamics.} To capture the runtime system dynamics of the real-world UAV testbed, we monitor key system-level metrics throughout an entire flight. As shown in Figure~\ref{fig:system dynamic}, we record the overall voltage, current, and the usage of CPU, GPU, and memory. We highlight four representative time points: $T_1$ (data collection starts), $T_2$ (UAV takes off), $T_3$ (model inference begins), and $T_4$ (model inference ends).
During $T_1$, memory usage increases due to buffering of multimodal data. However, after takeoff at $T_2$, memory usage drops as the radar's power supply becomes unstable due to shared load from UAV motors, leading to reduced and intermittent data acquisition. This reflects the impact of resource contention on sensing quality. Once inference starts at $T_3$, GPU usage spikes, significantly increasing overall current draw. This stage also has higher CPU utilization due to data preprocessing and pipeline management. At $T_4$, all resource usage and power consumption drop as the system enters idle state after landing.
These variations highlight the importance of designing adaptive inference systems that are robust to dynamic runtime conditions in mobile edge deployments.

\begin{figure*}
    \vspace{-2mm}
    
    \begin{minipage}{1\textwidth}
        \centering
               \setlength{\abovecaptionskip}{0.cm}
        \setlength{\belowcaptionskip}{0.cm}
        \includegraphics[width=1\linewidth]{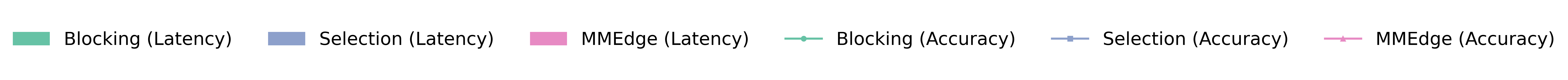}
    \end{minipage}
    \vspace{-3mm}
    
    \centering
    \begin{subfigure}[b]{0.24\textwidth}
        \centering
               \setlength{\abovecaptionskip}{0.cm}
        \setlength{\belowcaptionskip}{0.cm}
        \includegraphics[width=\linewidth]{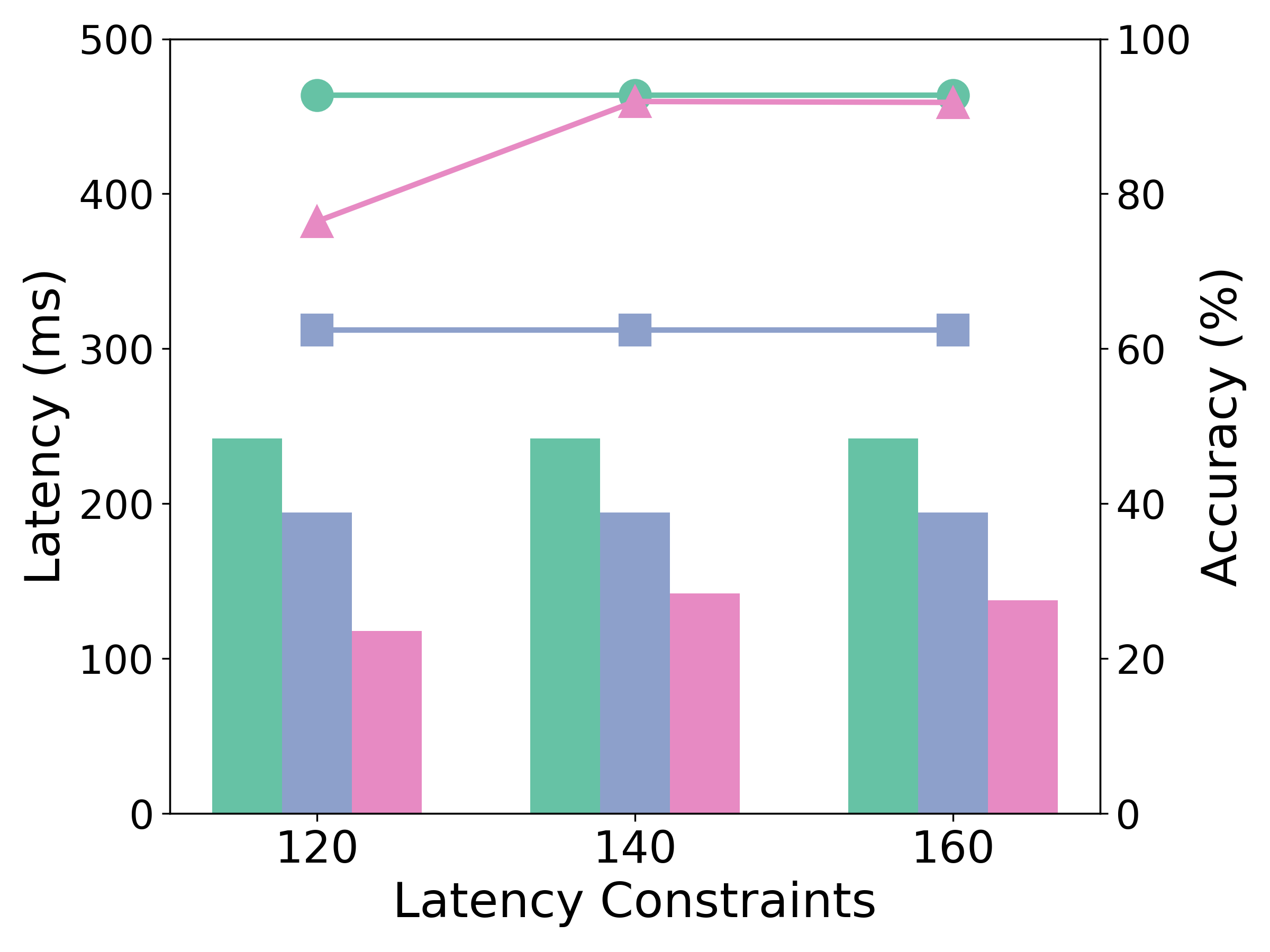}
        \caption{Different latency constraints.} 
        \label{fig:robustness latency}
        \vspace{-3mm}
    \end{subfigure}
    \begin{subfigure}[b]{0.24\textwidth}
        \centering
               \setlength{\abovecaptionskip}{0.cm}
        \setlength{\belowcaptionskip}{0.cm}
        \includegraphics[width=\linewidth]{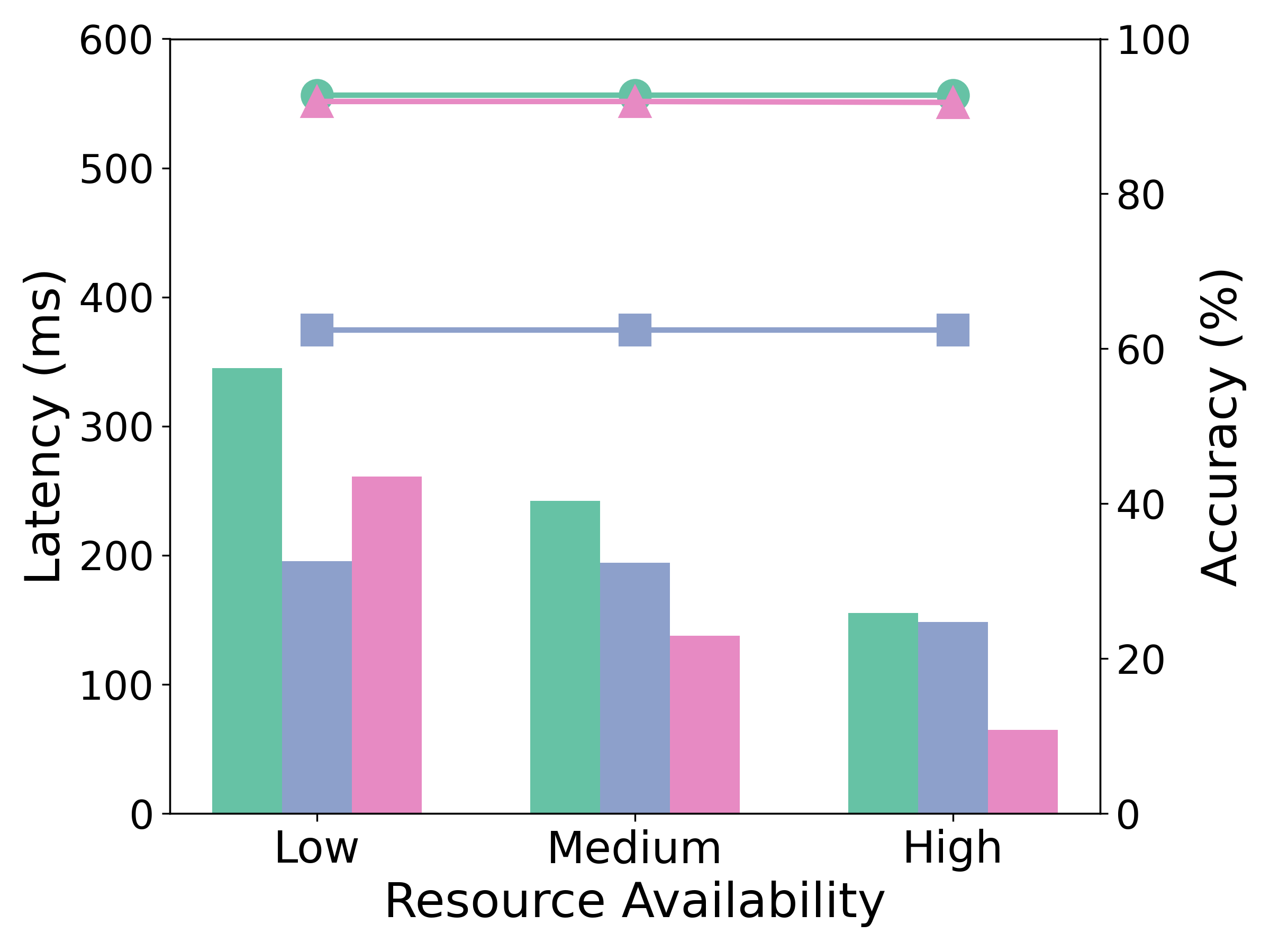}
        \caption{Different resource availability.}
        \label{fig:robustness resource}
        \vspace{-3mm}
    \end{subfigure}
    \begin{subfigure}[b]{0.24\textwidth}
        \centering
               \setlength{\abovecaptionskip}{0.cm}
        \setlength{\belowcaptionskip}{0.cm}
        \includegraphics[width=\linewidth]{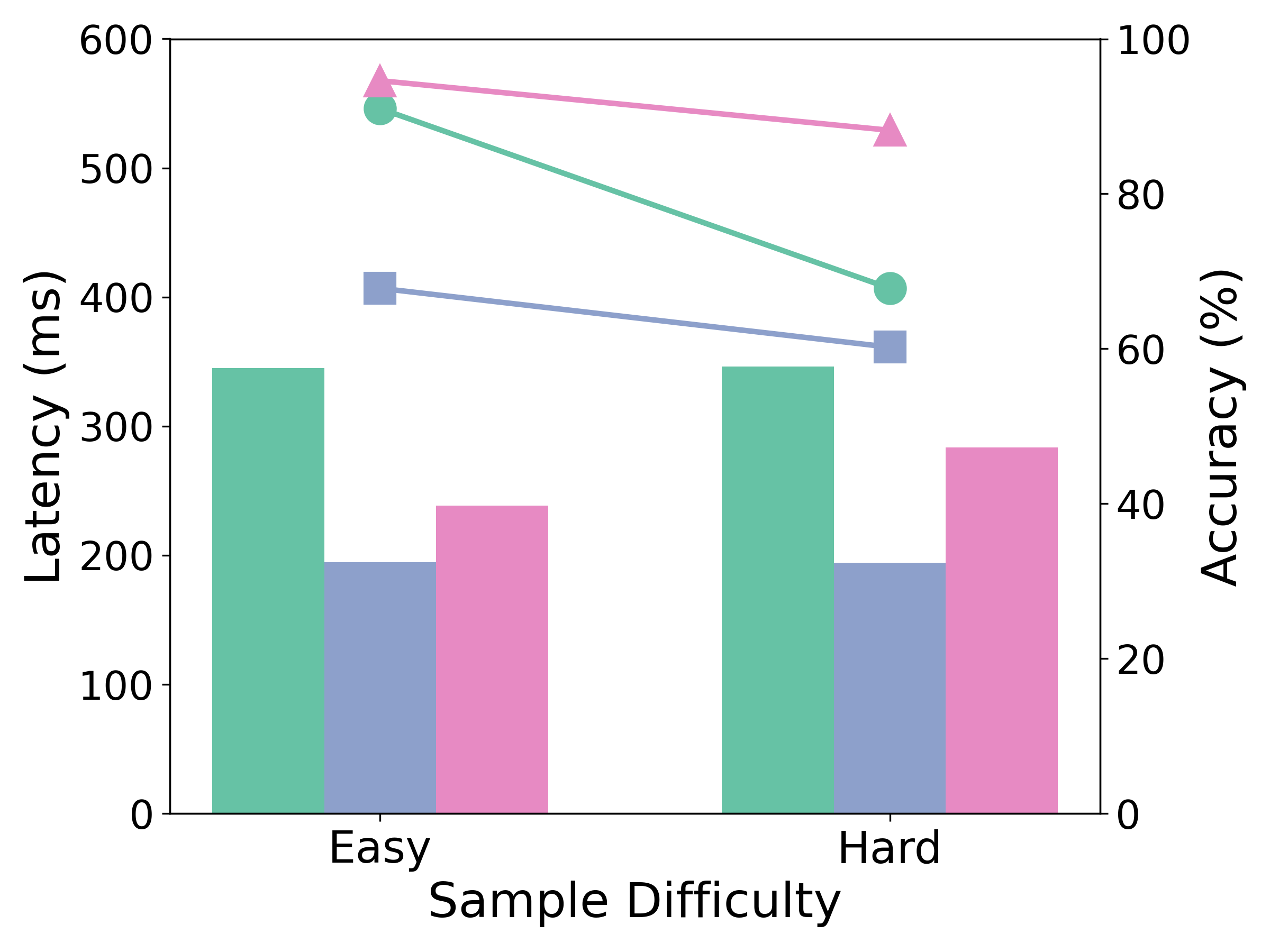}
        \caption{Diverse sample types.}
        \label{fig:robustness sample}
        \vspace{-3mm}
    \end{subfigure}
    \begin{subfigure}[b]{0.24\textwidth}
        \centering
               \setlength{\abovecaptionskip}{0.cm}
        \setlength{\belowcaptionskip}{0.cm}
        \includegraphics[width=\linewidth]{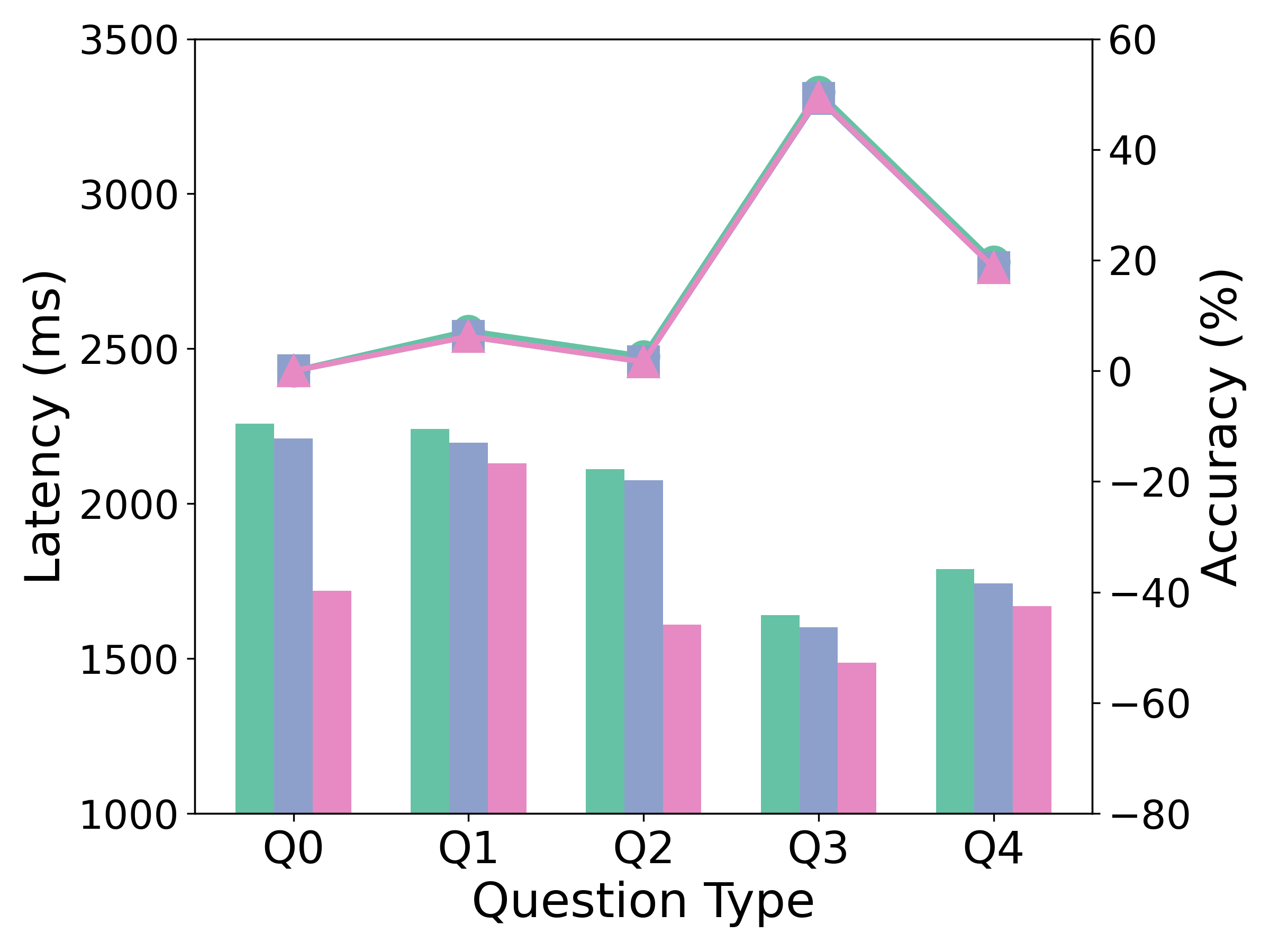}
        \caption{LLM decoding latency.}
        \label{fig:robustness decoding}
        \vspace{-3mm}
    \end{subfigure}
    
    \caption{Robustness of MMEdge under varying data and system conditions.}
    \vspace{-1.5em}
    \label{fig:robustness}
\end{figure*}

\vspace{-1em}
\subsection{Performance on Different Datasets}
Figure~\ref{fig:overall performance} shows the accuracy and end-to-end latency of various approaches across different multimodal datasets. The upper-left region of each plot indicates methods that achieve high accuracy with low latency, which is more favorable for real-time applications. 
First, we observe that each dataset exhibits different sensitivities to latency and accuracy trade-offs. For instance, in the NuScenes-Mini-QA dataset, latency varies significantly across methods, while accuracy remains relatively stable.  
In contrast, the LRW dataset shows a clearer trade-off between latency and accuracy, where reducing latency often comes at the cost of degraded performance. For evaluations the UAV testbed, detection performance (measured by mean IoU) also varies more substantially . 
Second, compared to baselines, MMEdge consistently achieves an optimal trade-off between high accuracy and low end-to-end latency. Blocking and Non-Blocking  methods deliver the highest accuracy but suffer from excessive latency, making them impractical for real-time use. Gating and Imputation reduce waiting time by bypassing slower modalities but still rely on holistic processing, failing to exploit the sensing interval and incurring significant latency. 
Pipelined Max and Min illustrate the performance bounds under fixed sensor and model configurations. In contrast, MMEdge dynamically adapts sensing and inference configurations based on different data inputs and real-time system dynamics, achieving competitive accuracy with significantly lower latency. This adaptability generalizes well across datasets with diverse modality and task characteristics, demonstrating MMEdge's effectiveness for low-latency on-device multimodal inference.
For example, on the UAV dataset, MMEdge achieves a mean IoU of 70.47\%, which is close to the Blocking baseline, while reducing end-to-end latency by over 80\%. On the LRW dataset, MMEdge compromises less than 3\% accuracy compared to the highest-performing baseline, but reduces latency to around 137 ms — the lowest among all methods. 
These results confirm MMEdge’s ability to retain high task performance under tight latency constraints across diverse multimodal scenarios.

\subsection{Robustness}

In this section, we evaluate the performance of MMEdge under varying data and system conditions to demonstrate its robustness. Specifically, we compare its accuracy and latency with representative data blocking and model selection baselines across different scenarios. The results are shown in Figure~\ref{fig:robustness}.


\subsubsection{Performance under Different Latency Budgets}
We evaluate MMEdge under different latency budgets \(T_{\text{max}} \in \{120, 140, 160\}\) ms to assess its adaptability. As shown in Figure~\ref{fig:robustness latency}, \name consistently maintains high accuracy (achieving 76.44\% at \(T_{\text{max}} = 120\) ms, and up to 91.92\% at \(T_{\text{max}} = 160\) ms), while keeping end-to-end latency within the specified limits. These results highlight MMEdge’s ability to dynamically adjust sensing and model configurations to meet strict latency requirements without sacrificing accuracy.

\subsubsection{Performance under Varying Resource Availability}
We also evaluate the robustness of MMEdge under different resource availability.
To simulate different levels of computational availability on Nvidia edge devices, we cap CPU usage at 100\% (high), 64\% (medium), and 50\% (low) using \texttt{cpulimit}~\cite{cpulimit}.
Figure~\ref{fig:robustness resource} shows that MMEdge consistently achieves low latency across all CPU levels while maintaining high accuracy, which demonstrates its ability to adaptively select sensing and model configurations in response to dynamic resource availability.

\subsubsection{Impact of Input Sample Complexity}
To assess the sensitivity of \name to input data dynamics, we categorize evaluation samples based on their difficulty level. Using the per-class accuracy distribution on the LRW dataset, we define the top 10\% as \textit{easy} and the bottom 10\% as \textit{hard} samples.
As shown in Figure~\ref{fig:robustness sample}, \name achieves 94.00\% accuracy with 293.1 ms latency on easy samples, and 87.40\% accuracy with 347.3 ms on hard samples. The increased latency for harder samples is primarily due to speculative skipping: 98\% of easy samples are skipped early, compared to only 90\% of hard samples, resulting in longer processing times.


\subsubsection{Impact of LLM Decoding Latency}
We evaluate MMEdge on the NuScenes-Mini-QA dataset under dynamic decoding latency introduced by large language models (LLMs) during answer generation. This dataset includes five question types: count, object, status, exist, and comparison. Unlike traditional classification models, the LLM generates free-form textual answers, resulting in variable and unpredictable decoding times. 
 To prompt the LLM, we use the format: \textit{Question: <question\_text> with data information: <prediction\_text>\textbackslash nAnswer:}. Accuracy is measured by checking whether the ground-truth answer appears in the generated output. Given a fixed overall task latency budget, the variability in LLM decoding time will affect the latency constraints passed to configuration optimizer in \name. As shown in Figure~\ref{fig:robustness decoding}, MMEdge effectively adapts to these fluctuations, maintaining high accuracy across all question types despite the unpredictable decoding delays.

\begin{figure}
    \raggedright   
    \setlength{\abovecaptionskip}{0.cm}
    \setlength{\belowcaptionskip}{-0.cm}
    \begin{subfigure}{0.5\linewidth}
            \setlength{\abovecaptionskip}{0.cm}
        \setlength{\belowcaptionskip}{0.cm}
        \raggedright   
        \includegraphics[width=1\linewidth]{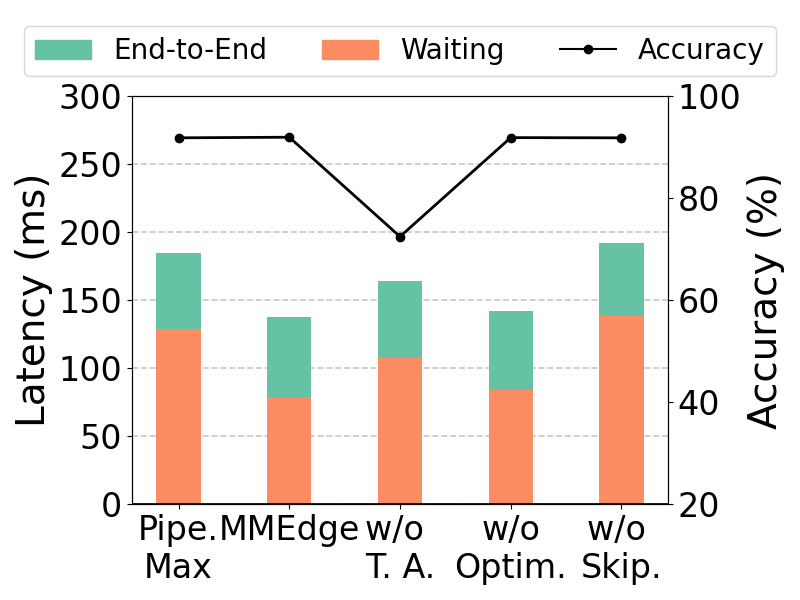}
        \caption{Ablation study.}
        \label{fig:ablation study}
    \end{subfigure}%
    \begin{subfigure}{0.5\linewidth}
        \centering
        \setlength{\abovecaptionskip}{0.cm}
        \setlength{\belowcaptionskip}{0.cm}
        \includegraphics[width=1\linewidth]{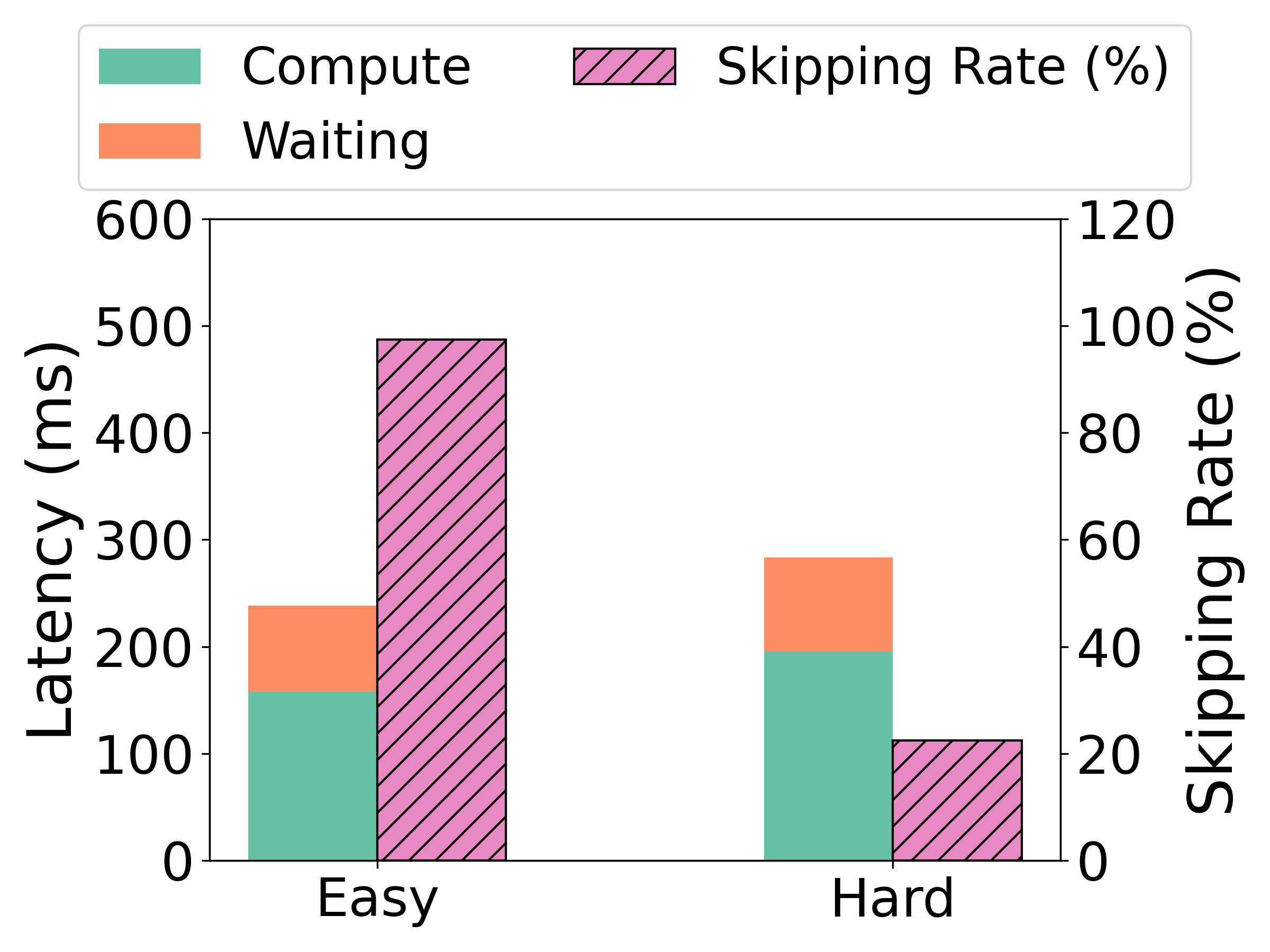}
        \caption{Modality skipping rates. 
        }
        \label{fig:effectiveness of skipping}
    \end{subfigure}
    \begin{subfigure}{1\linewidth}
        \centering
        \setlength{\abovecaptionskip}{0.cm}
        \setlength{\belowcaptionskip}{0.cm}
        \includegraphics[width=1\linewidth]{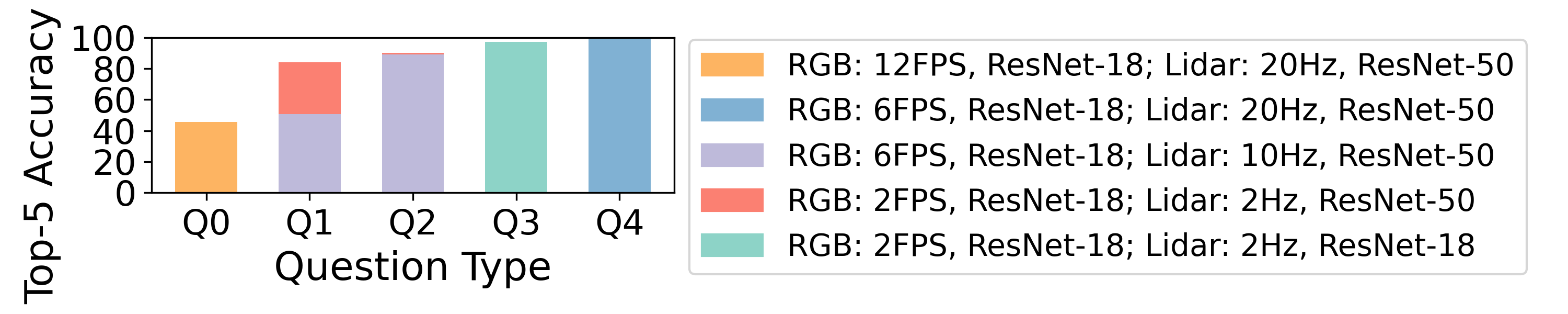}
        \caption{Selected configurations across different questions.}
        \label{fig:configuration distribution}
    \end{subfigure}
    \caption{Understanding MMEdge's Performance. 
    }
    \label{fig:ablation study and evaluation of accuracy predictor}
    \vspace{-2em}
\end{figure}

\subsection{Understanding MMEdge's Performance}

In this section, we perform ablation study and show the intermediate results to understand the effectiveness of \name. 

\subsubsection{Ablation study} \label{ablation_study} 
Figure~\ref{fig:ablation study} shows the performance of MMEdge on the LRW dataset when disabling different design components, including the temporal aggregation module (T.A.), adaptive multimodal configuration (optim.), and cross-modal speculative skipping (Skip.).
First, removing the temporal aggregation module results in a significant 19.5\% drop in accuracy, indicating its effectiveness in maintaining accuracy performance for the pipelined sensing and encoding framework.
When disabling speculative skipping module, the end-to-end latency increases from 137.67 ms to 192.03 ms, with the waiting time growing from 78.48 ms to 138.43 ms. This highlights that a significant portion of latency comes from waiting for slower modalities such as video.
Similarly, disabling adaptive configuration also leads to increased latency, highlighting their roles in optimizing performance under dynamic conditions.
On the LRW dataset, where the baseline accuracy is already high, the adaptive multimodal configuration primarily contributes to latency reduction rather than accuracy improvement, whereas in more diverse dataset (such as NuScenes-QA-Mini~\ref{fig:overall performance nuscenes}) it shows clearer benefits by enhancing the latency–accuracy trade-off.
These results collectively demonstrate that each component contributes to either accuracy or latency performance and is essential for achieving low-latency, high-accuracy inference.

\subsubsection{Effectiveness of Cross-Modal Speculative Skipping}. \label{effectiveness_skipping} 
To further analyze the effectiveness of the cross-modal speculative skipping module, we break down the results on the LRW dataset by sample difficulty. As shown in Figure~\ref{fig:effectiveness of skipping}, easy samples exhibit a higher skipping rate (76.2\%) compared to hard samples (22.4\%), resulting in lower latency (103.4 ms vs. 148.7 ms) while maintaining higher accuracy (94.8\% vs. 88.1\%).
These results demonstrate that the skipping module effectively identifies when early prediction is sufficient, selectively bypassing slower modalities without compromising accuracy.

\subsubsection{Effectiveness of Adaptive Multimodal Configuration} \label{effectiveness_configuration} 
We further evaluate the effectiveness of the adaptive multimodal configuration module using the NuScenes-QA-Mini dataset that has various types of questions.. First, we assess the performance of the accuracy predictor. Compared with the groundtruth model accuracy, it achieves a coefficient of determination ($R^2$ score) of 0.79 and a mean squared error (MSE) of 7.15, indicating reliable estimation of accuracy across different configuration combinations. Next, we profile classification accuracy by question type and the chose configuration by MMEdge, as shown in Figure~\ref{fig:configuration distribution}.
Across all question types, the difficulty gradually decreases from Q0 to Q4, as their Top 5 accuracy increases. Specifically, the questions in Q0 involve counting, which require temporal reasoning across multiple frames and are thus more error-prone, Q1 and Q2 involve object and status-related questions that depend on spatial recognition and are of moderate difficulty. Q3 (existence checking) and Q4 (comparison) questions are more straightforward, leading to higher accuracy.
We evaluate MMEdge under a 200 ms latency constraint to ensure that the optimizer can operate under a wide configuration space. 
As shown in Figure~\ref{fig:configuration distribution}, the optimizer selects more complex configurations for harder questions (e.g., Q0) and simpler ones for easier questions (e.g., Q4). This demonstrates that the adaptive multimodal configuration module effectively tailors sensing and model configurations based on task difficulty and latency constraints.

\begin{figure}
    \raggedright   
    \setlength{\abovecaptionskip}{0.cm}
    \setlength{\belowcaptionskip}{-0.cm}
    \begin{subfigure}{0.33\linewidth}
            \setlength{\abovecaptionskip}{0.cm}
        \setlength{\belowcaptionskip}{0.cm}
        \raggedright   
        \includegraphics[width=1\linewidth]{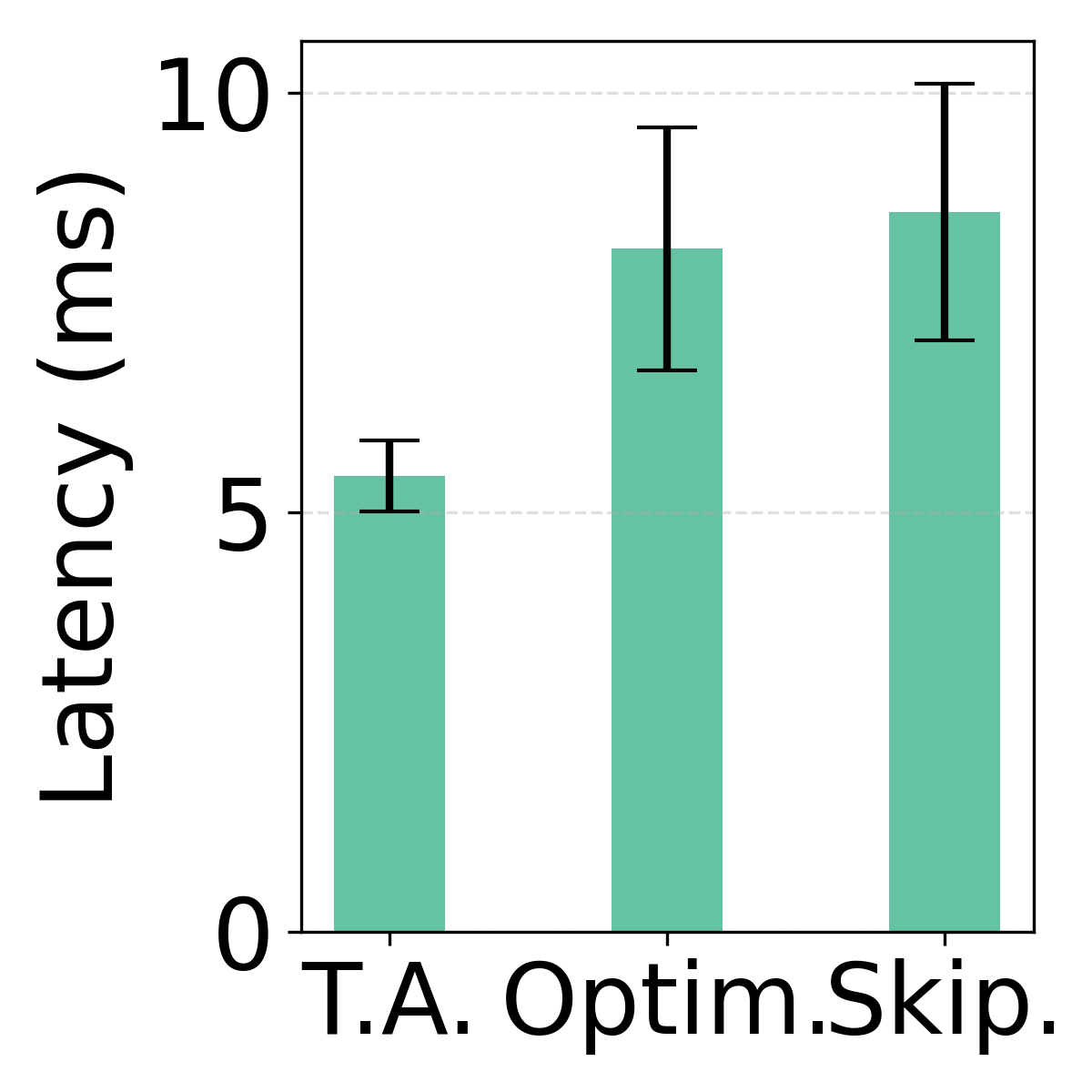}
        \caption{Latency.}
        \label{fig:latency overhead}
    \end{subfigure}%
    \begin{subfigure}{0.33\linewidth}
        \centering
        \setlength{\abovecaptionskip}{0.cm}
        \setlength{\belowcaptionskip}{0.cm}
        \includegraphics[width=1\linewidth]{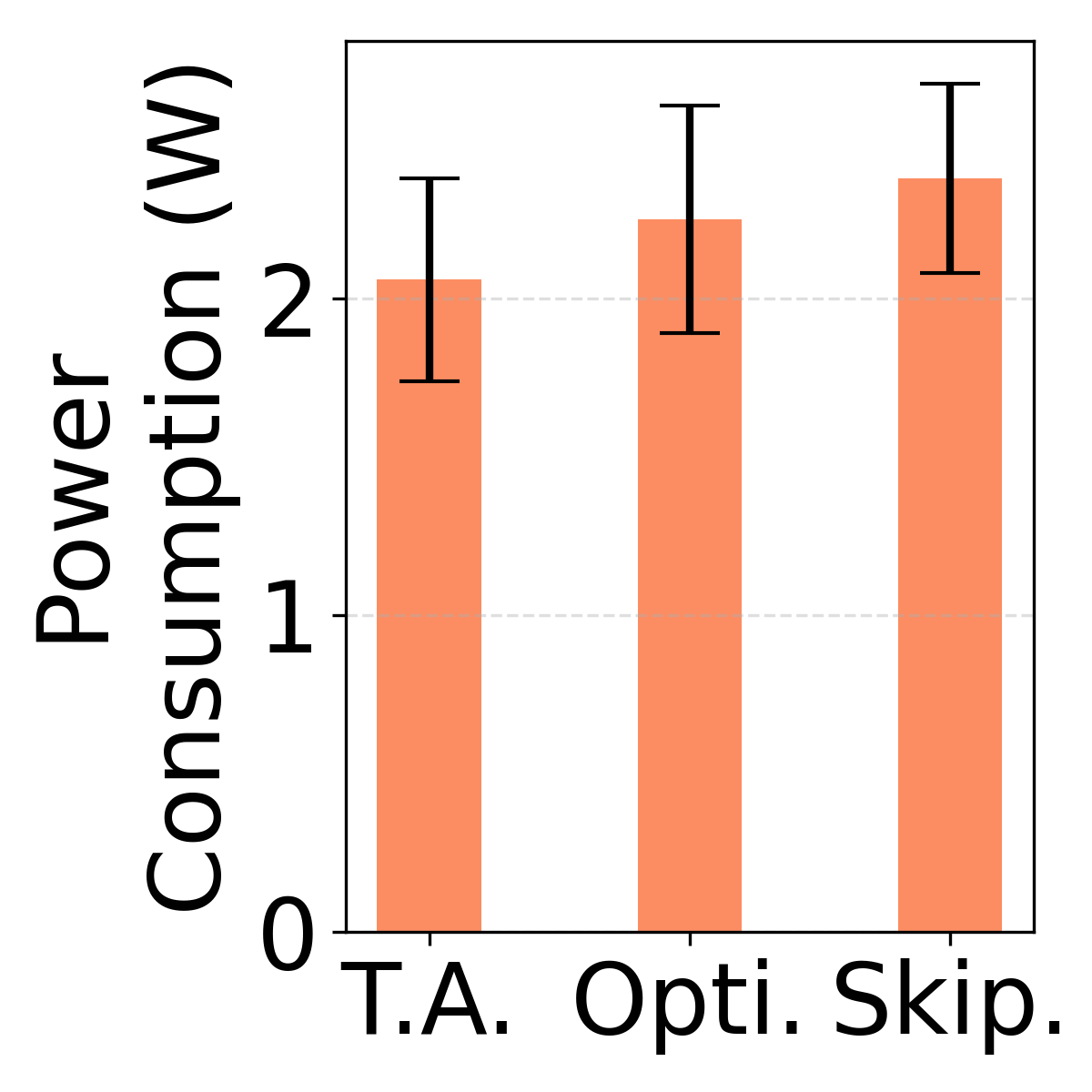}
        \caption{Power.}
        \label{fig:power overhead}
    \end{subfigure}
    \begin{subfigure}{0.33\linewidth}
        \centering
        \setlength{\abovecaptionskip}{0.cm}
        \setlength{\belowcaptionskip}{0.cm}
        \includegraphics[width=1\linewidth]{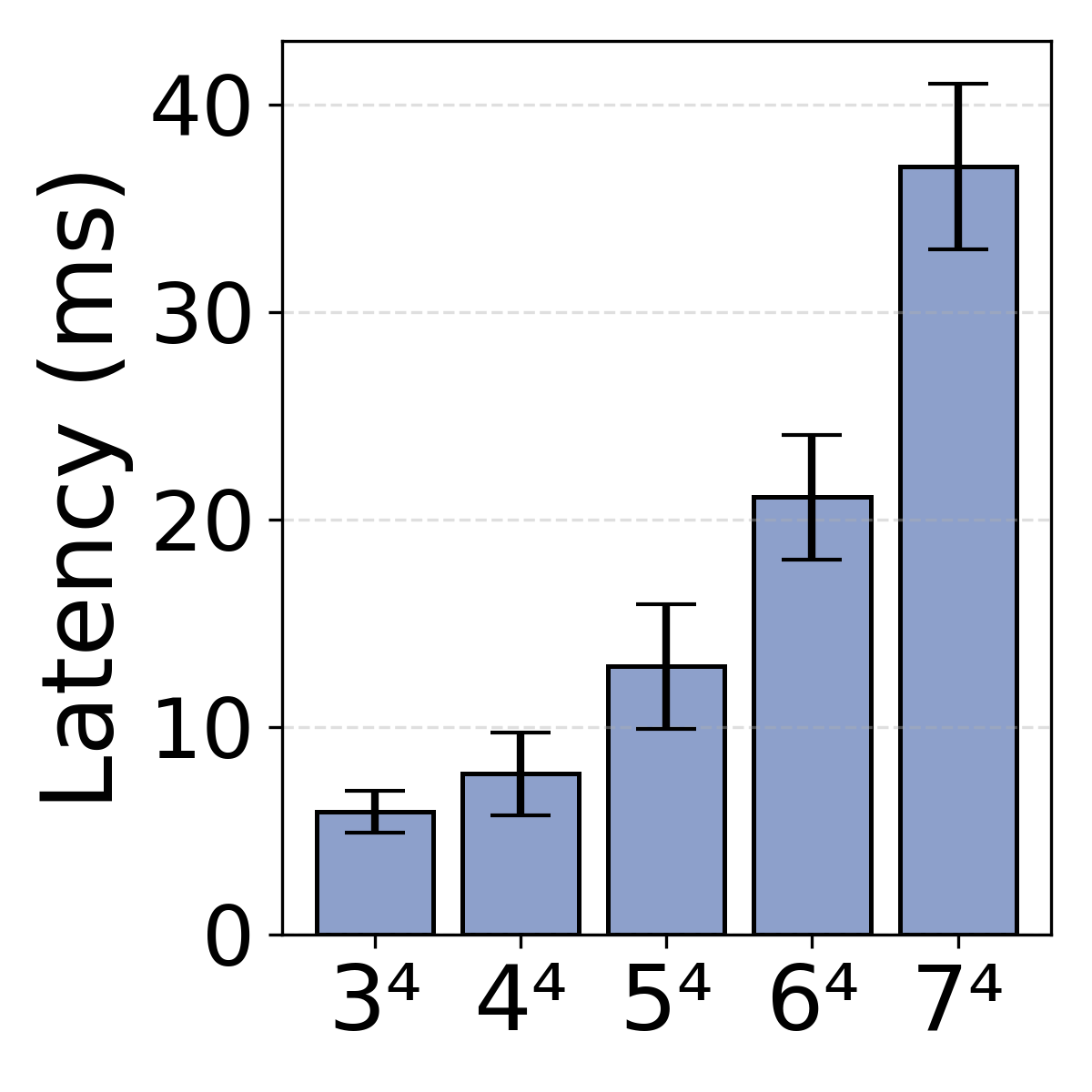}
        \caption{Scalability of Opti..}
        \label{fig:search space}
    \end{subfigure}
    \caption{Overhead of design modules. (T.A.: Temporal aggregation, Optim: Adaptive multimodal configuration optimizer, Skip.: Cross-modal speculative skipping.)} 
    \label{fig:overhead evaluation}
    \vspace{-2em}
\end{figure}

\subsection{Overhead of Design Modules}
We evaluate the overhead introduced by MMEdge’s core design modules: the temporal aggregation module within the pipelined sensing and encoding framework, the adaptive multimodal configuration optimizer, and the cross-modal speculative skipping module.

As shown in Figure~\ref{fig:latency overhead}, all components incur minimal latency overhead, with the highest being under 9 ms. Specifically, the temporal aggregation module adds 5.4 ms, the optimizer 8.2 ms, and the speculative skipping module 8.6 ms. These values are negligible compared to the overall system latency and are significantly outweighed by the latency reductions each module enables.
We also assess runtime power consumption introduced by our design modules, as shown in Figure~\ref{fig:power overhead}. The temporal aggregation module consumes 2.06 W on average, while the configuration optimizer and cross-modal skipping module increase power usage slightly to 2.25 W and 2.38 W, respectively. These results confirm that all modules operate within low energy budgets, and their marginal power overhead is well justified by the performance and latency gains they provide.
We also evaluate the overhead of the configuration optimizer by varying the search space, as shown in Fig.~\ref{fig:search space}.
Even when expanding the space from $3^4$ to $7^4$ combinations, the optimizer's latency increases only from 5.9 ms to 37.0 ms, which is negligible compared to the end-to-end latency (over 150 ms).
In practice, each modality typically has only a few configuration levels (e.g., camera: 10/20/30 FPS), enabling real-time configuration selection with minimal additional overhead.




\section{Discussion}

\noindent \textbf{Online profiling for \name.} \name uses offline profiling to build latency lookup tables for configuration selection, ensuring predictable optimization under stable conditions. However, real-world deployments often face resource fluctuations due to CPU contention, memory pressure, or temperature changes. In the future, we will explore integrateion of lightweight online profiling and periodic latency calibration, allowing \name to adapt to dynamic environments with minimal runtime overhead.

\noindent \textbf{Generalization ability of learning-based design modules.} \name’s learning-based components, such as the accuracy predictor and gating classifier, can model data-dependent behavior by training on diverse inputs. However, their accuracy may degrade under significant distribution shifts. To improve adaptability, we can incorporate online feedback calibration using recent inference results or apply lightweight unsupervised adaptation to these learning-based modules.

\noindent \textbf{Applicability to different multimodal fusion strategies and systems.} 
MMEdge currently adopts a simple fusion strategy aligned with its pipelined framework. However, as the design operates before fusion layers, it can be extended to support advanced strategies—such as attention- or gating-based fusion—for improved robustness to noisy or asynchronous inputs. Moreover, we will explore how MMEdge can be applied to more integrated platforms such as robotics, where tightly coupled hardware and control loops make runtime configuration changes challenging. To tackle this, we will extend MMEdge to support semi-flexible configurations, so that it can operate reliably while minimizing overhead.

\section{Conclusion}
In this paper, we present MMEdge, a new and efficient system for real-time on-device multimodal inference. MMEdge introduces a new pipelined sensing and encoding framework that decomposes inference into fine-grained units, allowing encoding computations to proceed concurrently with sensor data acquisition. It also integrates an adaptive multimodal configuration optimizer and a cross-modal speculative skipping mechanism to adapt to resource and data dynamics.
Extensive evaluations on two public multimodal datasets and a real-world UAV testbed show that MMEdge significantly reduces inference latency while preserving accuracy. 
In the future, we will leverage transfer learning techniques to generalizing MMEdge to unseen domains and devices with minimal supervision, and explore temporal redundancy across adjacent data samples to further optimize performance of MMEdge over time.



\begin{acks}
This work is supported by the Research Grants Council (RGC) of Hong Kong, China, under grant ECS 26200825, the HKUST - HKUST(GZ) Cross-campus Research Collaboration "1+1+1" Joint Funding Program under G\_2025\_052, and is partly funded by the HKUST Institute for Emerging Market Studies with support from EY, under grant IEMS25EG01.
\end{acks}

\bibliographystyle{ACM-Reference-Format}
\bibliography{reference}

\end{document}